\documentclass{fairmeta}
\usepackage{amsmath}
\usepackage{enumerate} 
\usepackage{bm}
\usepackage{floatflt}
\usepackage{algpseudocode}
\usepackage{amsfonts}
\usepackage{amsthm}
\usepackage{newtxtt}
\usepackage{colortbl} 
\usepackage{cleveref}
\usepackage{diagbox} 
\usepackage[utf8]{inputenc}
\usepackage{textgreek}
\usepackage{colortbl}
\usepackage{nicematrix}
\usepackage{makecell}
\usepackage{float}
\usepackage{arydshln}
\usepackage[frozencache,cachedir=.]{minted}
\usepackage{caption}
\usepackage{subcaption}
\usepackage{tcolorbox}
\usepackage{amssymb}
\usepackage{xspace}
\usepackage{wrapfig}
\usepackage{adjustbox}
\usepackage{tabularx}
\usepackage{booktabs}
\usepackage{mathtools}
\usepackage{wrapfig}
\usepackage{amssymb}
\usepackage{graphicx}

\usepackage{silence}
\makeatletter
\patchcmd{\wrong@fontshape}{\@gobbletwo}{}{}{}
\makeatother
\definecolor{upColor}{RGB}{17,138,21}
\definecolor{downColor}{RGB}{174,36,67}

\newtheorem{theorem}{Theorem}[]

\newtheorem{remark1}[theorem]{Remark}

\usepackage[linesnumbered,ruled,vlined]{algorithm2e}

\usepackage{graphicx}
\graphicspath{{./fig/}}

\usepackage[utf8]{inputenc}

\usepackage{booktabs}   
\usepackage{multirow}   
\usepackage{xcolor}     
\usepackage{graphicx}   
\usepackage{pifont}     
\usepackage{array}      
\usepackage{float}
\usepackage{setspace}

\newcolumntype{x}[1]{>{\centering\arraybackslash}p{#1}}

\title{TeleOCR: Navigating Document Parsing Across Digital and Camera-Captured Documents}

\author[1]{Peng Cai}
\author[1]{Zhaofan Zou$^*$}
\author[1]{Shifa Liu}
\author[1]{Yikun Wang}
\author[1]{Jiawei Tang}
\author[1]{Kaicheng Yang}
\author[1]{Meng Tong}
\author[1]{MingKun Jiang}
\author[1]{Zhongjiang He$^*$}
\author[1]{Hao Sun$^*$}
\affiliation[1]{China Telecom Artificial Intelligence Technology (Beijing) Co., Ltd.}

\metadata[$^*$Correspondence to]{Zhaofan Zou (\email{zouzhf41@chinatelecom.cn}), Hao Sun (\email{sun.010@163.com})}
\metadata[Model]{\url{https://huggingface.co/StarDoc-AI/TeleOCR}}
\metadata[Code]{\url{https://github.com/caipeng328/TeleOCR}}

\begin{document}

\abstract{
Document parsing aims to transform unstructured documents into structured and machine-readable representations. Recent advances in Vision-Language Models (VLMs) have significantly advanced document parsing. However, existing approaches still face two major challenges. First, decoupled VLM-based methods heavily rely on accurate layout analysis, where geometric distortions in camera-captured documents can introduce cascading errors. Second, although end-to-end VLM-based methods alleviate the dependence on explicit layout detection, they often suffer from redundant generation, hallucinations, and insufficient structural reasoning in high-resolution scenarios. To address these challenges, we propose TeleOCR, a unified framework for document parsing. TeleOCR introduces deformation-aware learning to incorporate geometric perception into VLMs and proposes an adaptive sampling mechanism for complex layout representation. Furthermore, a content-structure decoupled learning strategy is developed to explicitly model formula grammars and table structures, enabling more effective structured representation learning. Extensive experiments demonstrate that TeleOCR achieves state-of-the-art performance across diverse document parsing benchmarks. It obtains overall scores of 96.87, 88.53 and 78.41 on OmniDocBench v1.6, Wild-OmniDocBench, and PureDocBench, respectively, and ranks first in the ICDAR 2026 Sci-ImageMiner Challenge. These results validate the effectiveness and generalization capability of TeleOCR in complex document parsing scenarios.
}



\maketitle
\begin{figure*}[h]
\centering
\includegraphics[width=1\textwidth, trim=0mm 70mm 105mm 0mm, clip]{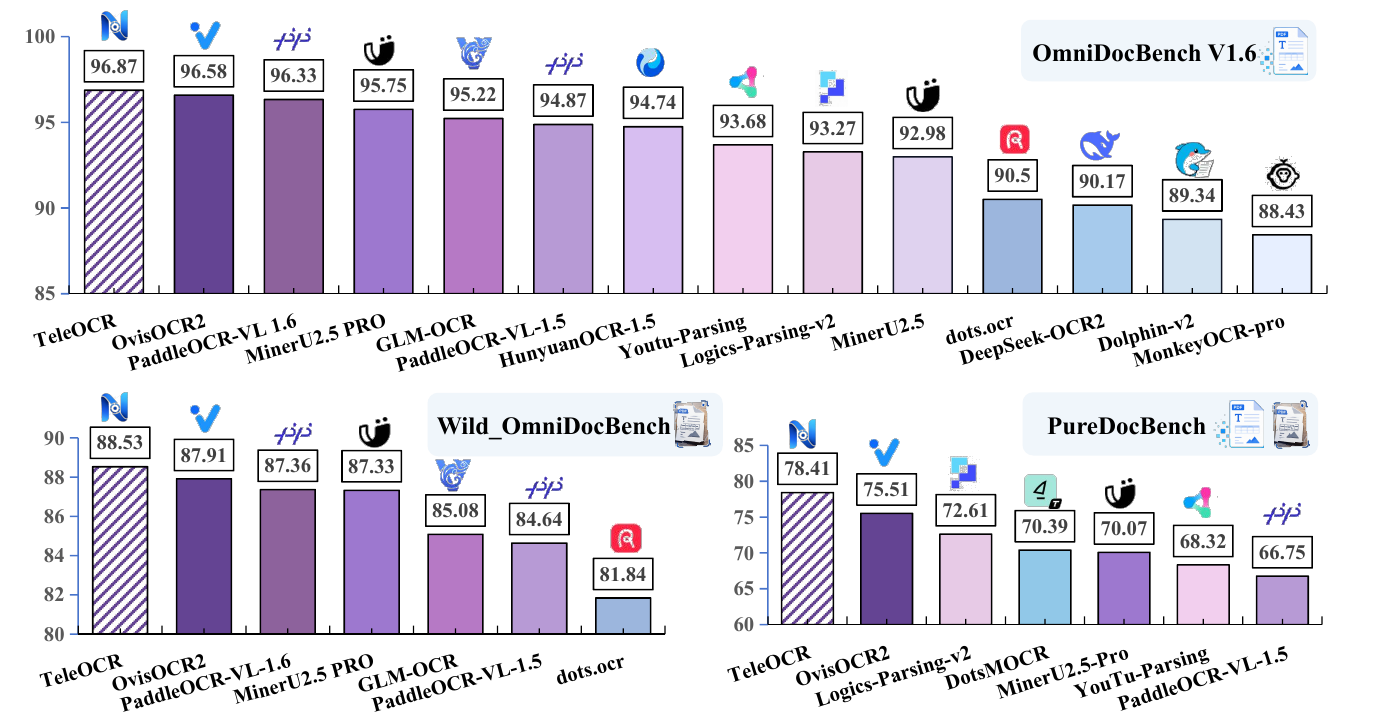} 
\caption{TeleOCR enables unified parsing of digital and camera-captured documents, achieving SOTA performance on OmniDocBench V1.6, Wild-OmniDocBench, and PureDocBench over end-to-end and decoupled approaches.}
\label{fig:framework}
\end{figure*}

\tableofcontents
\newpage

\section{Introduction}
Document parsing aims to transform unstructured documents into structured representations, such as Markdown, and serves as a fundamental component for constructing training data pipelines and Retrieval-Augmented Generation (RAG) systems~\cite{guo2025rag, wang2025bookrag}. Since parsing results directly affect downstream applications, accurate text recognition, layout understanding, and structural reconstruction are essential.

With increasing diversity in document acquisition conditions, document parsing systems need to handle both digital and camera-captured documents with complex degradations. To evaluate performance under these scenarios, several benchmarks have been introduced, including OmniDocBench V1.6~\cite{wang2026mineru2} for digital documents and Wild-OmniDocBench~\cite{li2026towards} for camera-captured documents. Correspondingly, existing VLM-based document parsing methods have developed into two paradigms: end-to-end and decoupled approaches. End-to-end methods are generally more robust to geometric distortions, whereas decoupled approaches perform better on high-resolution digital documents. However, decoupled approaches rely heavily on accurate layout analysis, where errors can propagate to subsequent parsing stages. To investigate this issue, we apply document dewarping preprocessing~\cite{cai2025forcennet} to Wild-OmniDocBench samples and find that removing geometric distortions alone substantially improves two-stage parsing performance.

Meanwhile, previous studies~\cite{zhong2025reading} indicate that documents with a high proportion of structured content generally exhibit higher prediction entropy, reflecting greater uncertainty during structured representation generation. In contrast, OCR tasks mainly rely on character-to-text mapping and therefore involve more stable generation processes. However, tasks such as table parsing and scientific figure-to-table conversion require models to not only understand visual content but also perform structural modeling and cross-modal transformation, which further increases the difficulty of learning and optimization.

Based on the above observations, TeleOCR aims to provide a unified solution for parsing both digital documents and camera-captured documents with distortions. To this end, TeleOCR introduces a global point-level and region-level deformation-aware learning strategy, which integrates document geometric rectification capabilities into VLMs. Combined with an adaptive sampling point mechanism, it replaces conventional detection paradigms with layout segmentation, enabling fine-grained document layout modeling. TeleOCR achieves an overall score of 88.53 on Wild OmniDocBench v1.5, outperforming most existing end-to-end document parsing methods.

Furthermore, highly structured tasks, such as formula parsing and table parsing, require simultaneous modeling of structural reasoning and content generation, increasing optimization complexity. To address this issue, TeleOCR proposes a content-structure decoupled learning strategy that explicitly separates structural prediction from content generation. Taking table parsing as an example, the model first predicts the table OTSL structure~\cite{lysak2023optimized} and then reconstructs cell contents based on the predicted structure, thereby explicitly modeling structural information and reducing optimization coupling. This strategy is also effective for scientific figure-to-table conversion by enhancing structural modeling capability, enabling TeleOCR to achieve first place in the ICDAR 2026 Sci-ImageMiner Challenge~\cite{ahmed2026icdar}.

The contributions are summarized as follows:

\begin{enumerate}
\item We propose TeleOCR, a unified document parsing framework that implicitly integrates document dewarping capabilities into VLMs through global point-level and region-level deformation-aware learning and an adaptive sampling point mechanism, enabling unified parsing of both digital and camera-captured documents.
\item We propose a content-structure decoupled learning strategy for highly structured document parsing tasks, which explicitly models formula grammars and table structures as intermediate reasoning processes. This strategy effectively reduces uncertainty in structure generation and significantly improves performance on formula parsing, table parsing, and scientific figure-to-table conversion tasks.
\item We conduct comprehensive evaluations on multiple public benchmarks and competitions. Experimental results demonstrate that TeleOCR achieves superior performance across digital document, camera-captured document, and scientific document parsing tasks. It obtains an overall score of 96.87 on OmniDocBench v1.6, an overall score of 88.53 on Wild OmniDocBench v1.5, and ranks first in the ICDAR 2026 Sci-ImageMiner Challenge, validating its effectiveness and generalization capability.
\end{enumerate}

\section{Related Work}
\label{sec:related_work}

\subsection{VLM-based Document Parsing Methods}
Existing document parsing methods based on Vision-Language Models (VLMs) can be broadly categorized into two groups: end-to-end VLM approaches and decoupled VLM approaches. End-to-end methods directly map document images into structured representations through unified vision-language modeling, avoiding error accumulation in traditional pipelines. Recently, OCR-oriented end-to-end models have attracted increasing attention. OvisOCR~\cite{lu2026ovisocr2} improves text and layout parsing in high-resolution documents by optimizing visual information interaction mechanisms. DeepSeek-OCR~\cite{wei2025deepseek} explores an OCR paradigm that integrates visual compression with language model reasoning, reducing the computational cost of long-document parsing. HunyuanOCR~\cite{hunyuanvisionteam2025hunyuanocrtechnicalreport} achieves unified multi-task modeling through a high-resolution vision encoder and a lightweight language model. In addition, Logics-Parsing~\cite{LogicsParsing} enhances complex layout understanding through layout-aware reinforcement learning. However, end-to-end methods typically suffer from high computational overhead in high-resolution scenarios and strong coupling between structural reasoning and content generation, limiting their scalability for complex document parsing.

In contrast, decoupled VLM methods combine the controllability of traditional pipelines with the semantic modeling capability of VLMs, and generally adopt a two-stage paradigm of "layout analysis followed by content parsing." Dolphin~\cite{feng2025dolphin} introduces an analyze-then-parse framework, where layout element sequences guide region-level content parsing. MonkeyOCR v1.5~\cite{zhang2025monkeyocrv15technicalreport} and GLM-OCR~\cite{glm_ocr} improve table recognition and OCR inference efficiency from the perspectives of visual consistency optimization and efficient decoding, respectively. Youtu-Parsing~\cite{YoutuParsing} further explores shared visual representations and parallel decoding mechanisms to reduce inference costs. For camera-captured document scenarios, PaddleOCR-VL-1.5~\cite{paddleocr_vl15} introduces multi-point bounding box modeling to handle physically degraded layouts, while PaddleOCR-VL-1.6~\cite{zhang2026paddleocr} and MinerU2.5-Pro~\cite{wang2026mineru2} improve parsing performance through data optimization and multi-model fusion, respectively. Nevertheless, existing decoupled approaches still heavily rely on accurate layout analysis. Geometric distortions can propagate errors through subsequent modules, limiting their effectiveness in complex camera-captured document scenarios. TeleOCR follows the decoupled VLM paradigm and further improves its capability by replacing conventional rectangular detection with layout segmentation, incorporating deformation-aware learning, and introducing a content-structure decoupled strategy. These designs enable unified parsing of both digital and camera-captured documents while enhancing performance on complex structured document understanding tasks.

\subsection{Document Rectification for Enhanced Parsing}
Document dewarping enhancement aims to correct perspective distortions and geometric deformations in camera-captured documents, thereby improving subsequent parsing performance. DDCP~\cite{xie2021document} predicts a fixed number of foreground control points and estimates backward mappings based on the correspondence between control points and reference points, enabling document dewarping. DocGeoNet~\cite{feng2022geometric} introduces segmentation supervision to encourage CNN-based text-line feature extractors to learn more discriminative geometric rectification features. RDGR~\cite{jiang2022revisiting} first detects text lines and boundary information, and then generates backward mappings with grid regularization to preserve document structural integrity during the dewarping process. ForCenNet~\cite{cai2025forcennet} further explicitly models document foreground regions and enhances the model's awareness of foreground geometric structures through curvature consistency loss and mask-guided mechanisms.Different from the above methods based on explicit geometric modeling, this paper integrates document rectification capability into a Vision-Language Model (VLM), enabling the model to directly learn deformation-related geometric control points. Meanwhile, region-level and global point-level deformation-aware mechanisms are designed to transform document rectification from an independent preprocessing module into an internal joint modeling capability of the VLM. This design improves the parsing performance of decoupled VLMs in camera-captured document scenarios.

\subsection{Vision-Language Model-based Segmentation}
Multimodal Large Language Model (MLLM)-based methods for region segmentation can be broadly categorized into two groups: one directly generates segmentation results through end-to-end sequence prediction, while the other introduces dedicated segmentation heads for task adaptation. SAM-based methods, such as SAM4MLLM~\cite{chen2024sam4mllm}, leverage external segmentation models to obtain high-quality masks, but introduce additional parameters and deployment overhead. Furthermore, SAM3~\cite{carion2025sam} extends foundation segmentation models to concept-prompted segmentation, object detection, and tracking tasks, further strengthening the external model paradigm. In contrast, SAM-free methods achieve region segmentation through lightweight dense prediction heads or unified autoregressive modeling, such as PerceptionGPT~\cite{pi2024perceptiongpt}, UFO~\cite{tang2026ufo}, and Qwen3-VL-Seg~\cite{yao2026qwen3}. For end-to-end sequence prediction, VistaLLM~\cite{pramanick2024jack} proposes a gradient-based dynamic sampling strategy to convert binary masks into point sequence representations. Considering the requirements of model unification and efficiency, this paper further reformulates conventional two-point layout analysis as a VLM-based sequence prediction task. A Curvature-Guided Douglas--Peucker Sampling (CGDP) method is proposed to adaptively distribute sampling points according to document geometric deformation characteristics, thereby enhancing the capability of VLMs to model complex layout structures.

\begin{figure*}[h]
\centering
\includegraphics[width=0.95\textwidth, trim=58mm 5mm 70mm 1mm, clip]{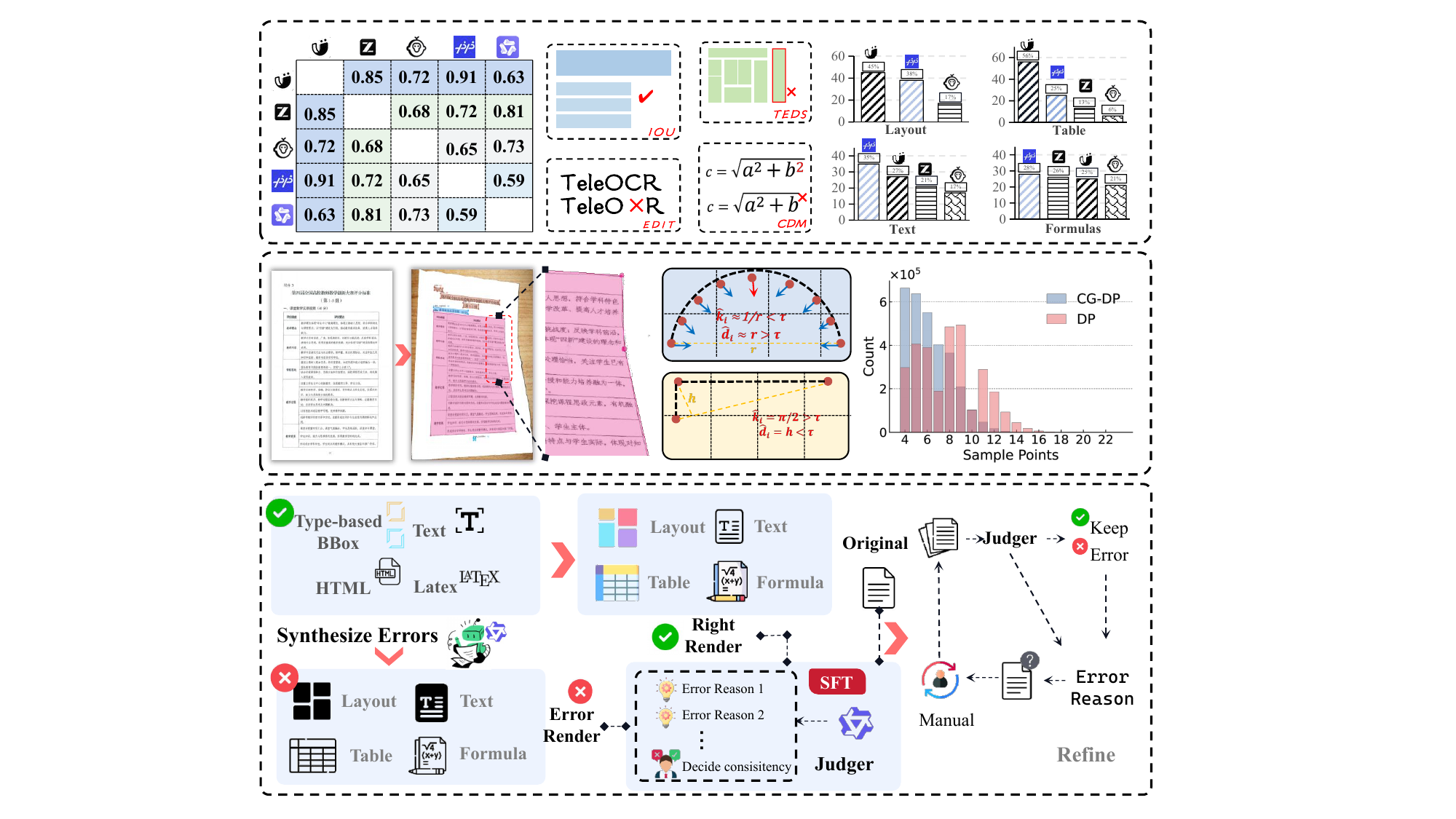} 
\caption{Overview of the TeleOCR data engineering pipeline. Large-scale document parsing data are constructed through three stages: (1) multi-node consensus voting for high-quality pseudo-label generation, (2) deformation-aware synthesis to bridge digital and camera-captured documents, and (3) image-to-image consistency evaluation with Self-Judgement VLM for automatic validation and refinement. The pipeline covers diverse document elements, including layouts, texts, tables, and formulas.}
\label{fig:framework}
\end{figure*}

\section{Data Engineering}
\label{sec:Data_Engineering}
High-quality and large-scale data are fundamental to building high-performance document parsing models. However, document parsing involves diverse structural information, including text, layouts, tables, and formulas. Relying on human experts for fine-grained annotation is not only costly but also difficult to maintain annotation consistency. To address this challenge, TeleOCR develops an automated data engineering pipeline that integrates data cleaning, data construction, and model-assisted validation. \textbf{In the first stage}, multi-node consistency voting aggregates predictions from heterogeneous models to select high-confidence pseudo-labels and reduce individual model biases. \textbf{In the second stage}, document deformation modeling and geometry-aware sampling strategies are introduced to construct diverse training data covering domain variations between digital and camera-captured documents. \textbf{In the third stage}, a self-evaluation model performs visual consistency verification on parsing results, enabling automatic pseudo-label filtering and error correction.


\subsection{Multi-node Consensus Voting}
Existing approaches mainly rely on a single model for pseudo-label generation. For instance, MinerU2.5 Pro~\cite{wang2026mineru2} adopts multi-model cross-validation to filter single-model outputs, while PaddleOCR-VL-1.5~\cite{paddleocr_vl15} exploits inference consistency across multiple runs of the same model for sample selection. However, these methods share a fundamental limitation: pseudo-labels are ultimately generated by a single model and are therefore bounded by its capability. Once the model suffers from systematic errors, these incorrect labels can be propagated into the training set, degrading subsequent optimization and final model performance.

To mitigate single-model bias, we propose a Multi-node Consensus Voting (MCV) strategy for reliable pseudo-label generation. Instead of relying on individual predictions, MCV leverages the consensus among multiple heterogeneous models to identify high-quality pseudo-labels. Given a set of $N$ heterogeneous models $\mathcal{M}=\{M_1,M_2,\cdots,M_N\}$, each model predicts the same sample $x$, producing a prediction set $\mathcal{Y}(x)=\{y_1,y_2,\cdots,y_N\}$, where $y_i=M_i(x)$.

\textbf{MCV is built upon the Consensus Hypothesis}: for heterogeneous models with different architectures and training strategies, predictions supported by multiple models are more likely to be reliable than individual predictions. Based on this hypothesis, MCV defines a pairwise consistency function $S(y_i,y_j)\in[0,1]$ to measure the agreement between any two predictions. The consistency metric is task-specific, including Intersection-over-Union ($IoU$) for layout detection, Edit Distance for text recognition, TEDS for table parsing, and CDM for formula parsing. The overall consensus score of model $M_i$ is then defined as:
\begin{equation}
 C_i=\frac{1}{N-1}\sum_{j\neq i}S(y_i,y_j)
\end{equation}
where $C_i$ measures the average agreement between prediction $y_i$ and the predictions from other models. The prediction with the highest consensus score is selected as the pseudo-label:
\begin{equation}
\hat{y}=y_{k},\quad k=\arg\max_i C_i .
\end{equation}
To further improve pseudo-label quality, a consensus threshold $\tau$ is introduced. If $\max_i C_i\geq\tau$, the corresponding prediction is accepted as a high-confidence pseudo-label and added to the training set. Otherwise, the sample is considered uncertain due to substantial model disagreement and is forwarded to subsequent automatic correction or human verification modules.

\subsection{Unifying Digital and Camera-Captured Documents}
According to the acquisition process, documents can be categorized into digital documents and camera-captured documents. Compared with digital documents, camera-captured documents often suffer from various degradations, such as geometric distortions, shadows, and motion blur, with geometric distortion being a primary factor affecting parsing performance. In particular, existing two-stage document parsing frameworks typically rely on the assumption of regular rectangular regions for layout detection. When documents are curved or folded, this assumption no longer holds, resulting in degraded layout detection and subsequent parsing performance. To validate this observation, we first apply a document dewarping model to preprocess camera-captured documents in Wild-OmniDocBench~\cite{li2026towards}. Experimental results show that eliminating geometric distortions alone substantially improves the performance of two-stage parsing models. This finding suggests that deformation awareness is a critical bridge between digital and camera-captured documents. Motivated by this, we incorporate document deformation modeling into a unified document parsing framework, enabling explicit geometric perception without requiring an additional dewarping module.

\subsubsection{Region-level and Point-level Deformation Awareness}

Based on the high-consistency samples selected by MCV, we construct two types of deformation supervision signals: region-level and point-level representations. For the region-level representation, $N$ boundary points are uniformly sampled in a clockwise order along the undistorted layout boundary $L_{bbox}$ and serialized as $[R_{x1},R_{y1},\cdots,R_{xN},R_{yN}]$. For the point-level representation, $M\times M$ control points $P$ are uniformly sampled on the undistorted document plane.

Subsequently, we adopt the document deformation generation strategy of ForCenNet~\cite{cai2025forcennet} to synthesize corresponding camera-captured document samples. Specifically, the original backward mapping $BM$ is obtained from Doc3D~\cite{das2019dewarpnet}, from which the forward mapping $FM$ is derived. The generated $FM$ is applied to the undistorted image, region boundary points $R$, and control points $P$, respectively, producing distorted document images with the corresponding warped boundaries $R_w$ and control points $P_w$.

During the early training stage, TeleOCR learns the distorted control points $P_w$ under explicit geometric supervision to capture document deformation patterns. Existing document dewarping methods typically represent the backward mapping field ($BM$) through dense control points $P_w$. Subsequently, we introduce distorted region boundary points $R_w$ to replace conventional rectangular bounding boxes and reformulate layout detection as a boundary point prediction task. This design removes the reliance of two-stage layout detection frameworks on regular rectangular assumptions and improves the model’s ability to represent complex layouts in camera-captured documents.

\subsubsection{Curvature-Guided Douglas--Peucker Sampling}
The aforementioned region-level representation relies on uniform boundary sampling, which assumes equal importance across all boundary locations. However, document boundaries often exhibit varying geometric complexity: flat regions may contain redundant samples, whereas high-curvature areas, such as corners and folds, may be under-sampled. This imbalance can lead to the loss of critical geometric details and limit the representation capacity of contour modeling.

To investigate sampling requirements under different deformation patterns, we categorize document deformations into two types: bending and creasing. Bending refers to large-scale continuous deformation with substantial global displacement, which can be effectively characterized by the Douglas--Peucker ($DP$) distance~\cite{hershberger1992speeding}. In contrast, creasing involves local directional discontinuities with high curvature. Due to its limited spatial extent, creasing regions may not generate sufficiently large $DP$ errors. Therefore, $DP$ mainly captures region-level geometric deviations, while curvature provides a more suitable measure of point-level local structural importance. For smooth curves, the DP distance approximately satisfies $d\approx\frac{\kappa L^{2}}{8}$, indicating that DP distance is jointly determined by local curvature and region scale. Based on this observation, we propose \textbf{Curvature-Guided Douglas--Peucker Sampling (CGDP)}, which adaptively adjusts the importance of $DP$ points through curvature-aware modulation:

\begin{equation}
S_i=d_i(1+\lambda\hat{\kappa}_i),
\end{equation}

where $\hat{\kappa}_i$ denotes the normalized local curvature. When the curvature is low, CGDP reduces to the standard DP algorithm. As curvature increases, points with prominent local structures receive higher sampling priority. When $\max_i S_i>\tau$, the corresponding point is selected as a new recursive node. By integrating the global contour preservation of DP with the local structural awareness of curvature, CGDP preserves critical geometric details, such as creases and sharp corners, under a limited sampling budget, resulting in more accurate document boundary representations.

\subsection{Self-Judgement VLM}
Although multi-model voting produces highly consistent pseudo-labels, they may still contain sample bias and residual errors. To address this limitation, we introduce a self-judgement model that renders structured predictions into visual representations and performs intra-modal consistency evaluation against the original document images. This design transforms conventional cross-modal image-text verification into a more stable image-to-image consistency assessment. Different from MinerU 2.5 Pro~\cite{wang2026mineru2}, which directly uses a general-purpose VLM to evaluate the consistency between original images and predicted results, we investigate the zero-shot verification capability of Qwen3-VL-235B~\cite{yang2025qwen3} on a self-constructed high-quality benchmark. The results show that its recall remains below $40\%$, demonstrating that general-purpose VLMs are insufficient for reliable unsupervised pseudo-label verification. Therefore, we convert four types of prediction outputs into unified visual representations and align them with the original document images:

\begin{enumerate}
\item \textbf{Layout}: Predicted bounding boxes, categories, and orientation information are projected onto a page canvas to reconstruct the overall document layout;
\item \textbf{Text}: Text content is normalized and reorganized into paragraphs, followed by region-aware re-layout to preserve the original textual structure;
\item \textbf{Table}: Table structures are converted into HTML representations with row-column relationships and merged-cell information, and then rendered into images;
\item \textbf{Formula}: LaTeX sequences are normalized and rendered into corresponding formula images.
\end{enumerate}


Given high-consistency pseudo-labels $\hat{y}$, erroneous predictions $\pi(\hat{y})$ are synthesized through rule-based and LLM-guided perturbations. These predictions are rendered into $\hat{I}_{\text{right}}$ and $\hat{I}_{\text{bad}}$, respectively, forming positive and negative pairs of "original image--correct rendering" and "original image--incorrect rendering" to train the self-judgement model. Specifically, layout perturbations include region-level errors (e.g., missing, displacement, and overlap) and structural-level errors (e.g., merging, splitting, and disorder). Text perturbations involve character-level corruption, text omission, and category misclassification. Table perturbations cover row-column structure errors, cell relationship errors, and content recognition errors. Formula perturbations include syntax errors, structural omissions, and type misclassification. Rule-based perturbations generate explicit and controllable error patterns, whereas LLM-guided perturbations simulate complex errors that require semantic understanding.
Based on the above image-to-image paired data, this paper performs supervised fine-tuning on Qwen2.5-VL-7B-Instruct~\cite{bai2025qwen25vltechnicalreport} to obtain the self-judgement model $J_{\theta}$. Given the original image $I(x)$ and the rendered result $\hat{I}_y$, the model analyzes their visual differences according to predefined checking criteria $\mathcal{E}=\{e_k\}_{k=1}^{K}$ :
\begin{equation}
r_k=M(I(x),\hat{I}_{y},e_k)
\end{equation}
The individual analysis results are aggregated into a structured reasoning sequence $R=(r_1,\ldots,r_K)$, which produces the final consistency decision. The trained $J_{\theta}$ can automatically filter pseudo-labels, identify error types, and generate error explanations, enabling self-correction of the data construction process. Low-confidence samples are further transferred to a human verification process.

\section{Progressive Training}
\begin{figure*}[!t]
\centering
\includegraphics[width=1\textwidth, trim=5mm 15mm 80mm 0mm, clip]{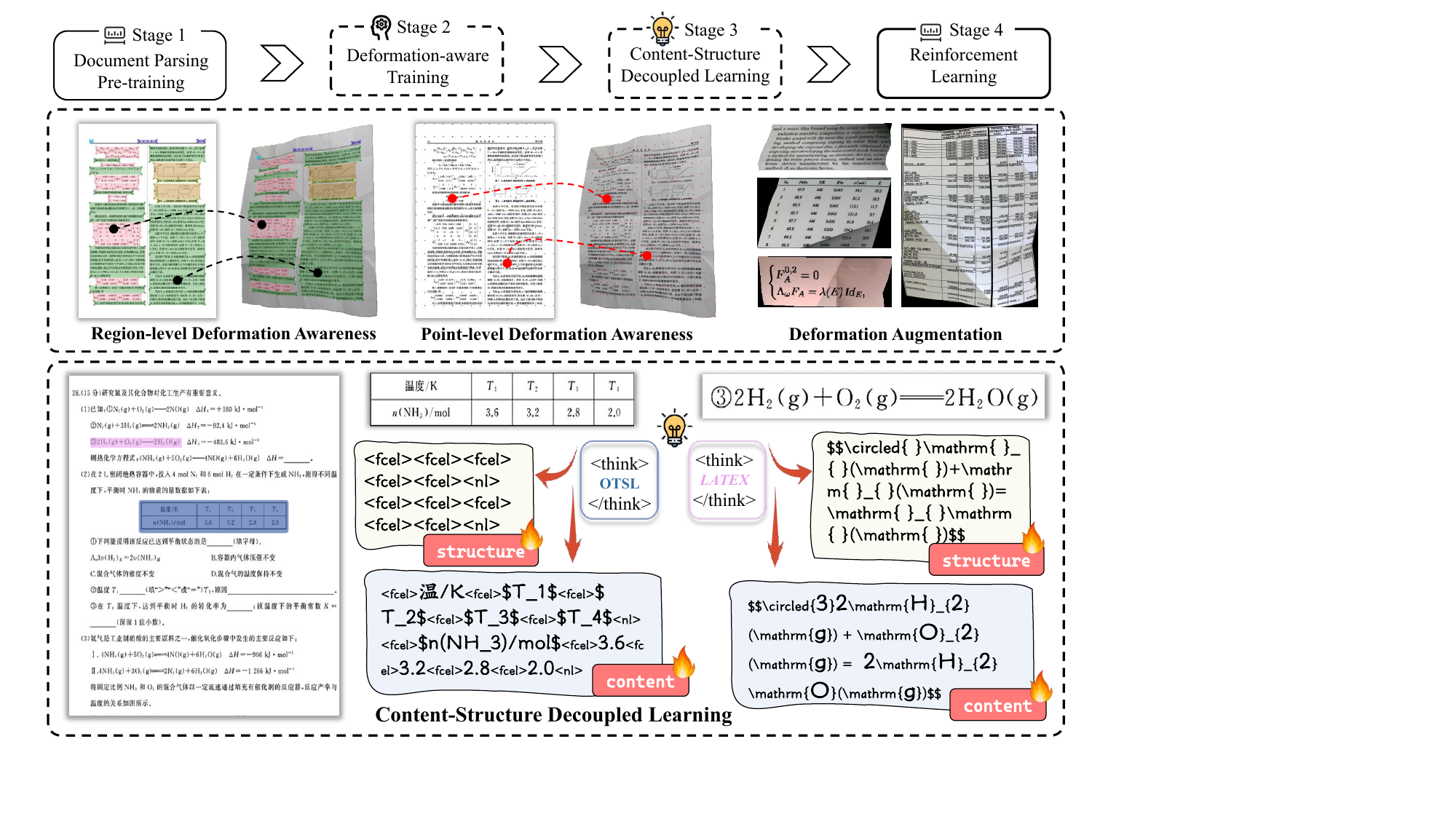} 
\caption{Overview of the key training strategies. TeleOCR enhances document parsing through two strategies: \textbf{Deformation-aware training}, which incorporates region-level and point-level deformation modeling with deformation augmentation for robust parsing of digital and camera-captured documents; and \textbf{Content-structure decoupled learning}, which enables unified structured parsing of tables and formulas.}
\label{fig:framework}
\end{figure*}

To facilitate reproducibility, we provide detailed descriptions of the TeleOCR architecture and training pipeline, and release the complete implementation and model configurations built upon community models. TeleOCR contains approximately 1.2B parameters and consists of a vision encoder inherited from Qwen2.5-VL~\cite{bai2025qwen25vltechnicalreport}, a Qwen3-0.6B~\cite{yang2025qwen3} language model, and an Aligner trained from scratch. The Aligner adopts a standard multi-layer perceptron (MLP) architecture to align visual and language representations.

Based on this unified architecture, TeleOCR adopts a four-stage progressive training strategy that gradually equips the model with document parsing capabilities, from basic visual perception to complex structural understanding. Each stage targets a specific optimization objective. \textbf{Stage 1} performs vision-language alignment to establish fundamental OCR recognition and layout understanding capabilities. \textbf{Stage 2} introduces region-level deformation-aware supervision to improve the model's ability to capture geometric distortions in camera-captured documents. \textbf{Stage 3} adopts a content-structure decoupled learning strategy to enable unified modeling of diverse document elements, including text, layouts, tables, and formulas. \textbf{Stage 4} further optimizes output quality through reinforcement learning, enhancing parsing performance in complex real-world scenarios.

\subsection{Stage 1: Document Parsing Pre-training}
The first stage focuses on aligning the vision encoder with the language model to establish a unified vision-language representation for subsequent document parsing learning. To provide TeleOCR with fundamental OCR recognition and layout understanding capabilities, we perform pre-training on visual question answering (VQA) data. During this stage, only the two-layer MLPs in the Patch Merger module and the vision encoder are optimized, while the language model remains frozen. The training data consists primarily of image captioning data, image-text interleaved data, vision-language alignment data, and OCR data.

\textbf{Training Configuration.} The model is trained for one epoch with a batch size of 256. The learning rates are set to $1\times10^{-3}$ for the MLP layers and $1\times10^{-4}$ for the vision encoder, while the language model parameters are kept frozen.

\subsection{Stage 2: Deformation-aware Training}
Stage 1 primarily establishes fundamental OCR recognition capabilities, while the ability to jointly model the layouts of digital and camera-captured documents remains limited. To enable the Vision-Language Model to explicitly capture geometric deformation patterns in camera-captured documents, we introduce a point-level deformation-aware training strategy. Specifically, the original deformation field is downsampled into $N^2=1,024$ control points, whereas dedicated document dewarping models (e.g., ForCenNet~\cite{cai2025forcennet}) typically employ 82,944 control points for dense deformation modeling. TeleOCR is then trained to directly predict the coordinates of the downsampled control points $P_i$, enabling document deformation modeling with substantially lower representation complexity.

\textbf{Training Data.} The training data in this stage consists of two components. \textbf{(1)} We collect 4M digital document layout samples and 2M synthetic camera-captured document samples. The synthetic samples include 1.2M region-level deformation samples and 0.8M global point-level deformation samples, constructed from layout coordinates $L_{bbox}$, global deformation control points $P_i$, and region boundary points $R_i$ generated by the CGDP sampling strategy. These annotations are selected using the MCV strategy. \textbf{(2)} Based on document parsing data refined by the data engine, approximately 120K high-quality samples are augmented with camera-captured styles to simulate realistic document degradations, including distortion, shadows, and blur. This augmentation further improves the model's ability to capture the distribution of camera-captured documents. The objective of this stage is to enhance deformation-aware representation learning for camera-captured documents.

\textbf{Training Configuration.} This stage adopts full-parameter fine-tuning. The learning rates are set to $1\times10^{-5}$ for the language model and $1\times10^{-6}$ for the vision encoder. The model is trained for one epoch with a batch size of 128.

\subsection{Stage 3: Content-Structure Decoupled Learning}

In document parsing tasks, formulas, tables, and scientific figures are typically converted into structured representations, such as LaTeX and OTSL. Unlike optical symbol recognition, which mainly focuses on local character-level mapping, structured representation generation requires stronger global semantic understanding and structural relationship modeling. After establishing fundamental OCR recognition capabilities in Stage 1, TeleOCR introduces a unified content-structure decoupled learning strategy to support diverse structured parsing tasks, including formula parsing, table parsing, and scientific figure-to-table conversion.

Specifically, for formula parsing, we construct a syntax token library and automatically extract syntax structure labels from LaTeX annotations through regular expression matching and syntax verification. This allows the model to separately learn formula content and grammatical structures. For table parsing, we preserve the standard OTSL syntax tokens while removing all cell contents, enabling the model to focus on table topology and cell merging relationships. This unified content-structure learning strategy improves the model’s ability to perform structured representation modeling, particularly for scientific figure-to-table conversion. In the ICDAR 2026 Sci-ImageMiner Challenge, incorporating structural learning achieves competitive TEDS performance compared with other participating teams.

\subsection{Stage 4: Reinforcement Learning}
After the first three training stages, TeleOCR acquires strong capabilities in document recognition and structured parsing. However, supervised fine-tuning with token-level cross-entropy loss does not directly optimize task-level objectives for text, table, and formula parsing. To further improve performance on downstream parsing tasks, we introduce Group Relative Policy Optimization (GRPO)-based reinforcement learning~\cite{shao2024deepseekmath} on the Stage 3 model.

Since different document elements have distinct output formats and evaluation criteria, we design task-specific verifiable reward functions for different parsing tasks. The unified formulation is defined as:

\begin{equation}
R_{\mathrm{task}}(y,\hat{y})=
\begin{cases}
1-\operatorname{NED}(y,\hat{y}), & Task=\mathrm{Text},\\
\operatorname{TEDS}(y,\hat{y}), & Task=\mathrm{Table},\\
\operatorname{CDM}(y,\hat{y}), & Task=\mathrm{Formula},
\end{cases}
\end{equation}

where $y$ and $\hat{y}$ denote the ground truth and model prediction, respectively. For text, table, and formula parsing, normalized edit similarity (NED), Tree Edit Distance-based Similarity (TEDS), and the formula structure matching metric (CDM) are adopted as task-specific rewards. All rewards are normalized to the range of $[0,1]$, where higher values indicate stronger consistency between predictions and ground truth.

\section{Experimental Evaluation}
To comprehensively evaluate TeleOCR, we conduct extensive experiments on multiple public document parsing benchmarks, including \textbf{OmniDocBench v1.6}~\cite{wang2026mineru2}, which covers 10 document types, 5 layout categories, and 5 languages; \textbf{Wild OmniDocBench v1.5}~\cite{li2026towards}, which evaluates robustness on real-world captured documents; and \textbf{PureDocBench}~\cite{li2026far}, which includes three document categories: Clean, Digital-Degraded, and Real-Degraded. We further report results on the \textbf{ICDAR 2026 Sci-ImageMiner} scientific figure-to-table conversion task to evaluate the generalization capability of TeleOCR in complex document understanding and parsing scenarios.

\subsection{Digital Document Parsing}

\begin{table*}[t]
  \centering
    \caption{Performance comparison of document parsing methods on OmniDocBench~v1.6 Full for text, formula, table, and reading order extraction.}
  \resizebox{1\textwidth}{!}{
    \begin{tabular}{c|l l|c|c c c c c}
    \toprule
    \textbf{Model Type} & \textbf{Methods} & \textbf{Param} & \textbf{Overall}$\uparrow$ & \textbf{Text\textsuperscript{Edit}}$\downarrow$ & \textbf{Formula\textsuperscript{CDM}}$\uparrow$ & \textbf{Table\textsuperscript{TEDS}}$\uparrow$ & \textbf{Table\textsuperscript{TEDS-S}}$\uparrow$ & \textbf{Read Order\textsuperscript{Edit}}$\downarrow$ \\
    \midrule
    \multirow{15}{*}{\makecell{\textbf{Specialized}\\\textbf{VLMs}}} 

    & \cellcolor{gray!20}\textbf{TeleOCR} & \cellcolor{gray!20}1.2B & \cellcolor{gray!20}\textbf{96.87} & \cellcolor{gray!20}\underline{0.027} & \cellcolor{gray!20}96.36 & \cellcolor{gray!20}\textbf{97.05} & \cellcolor{gray!20}\textbf{98.52} & \cellcolor{gray!20}\underline{0.122} \\
    
    & OvisOCR2~\cite{lu2026ovisocr2} & 0.8B & \underline{96.58} & \textbf{0.025} & \textbf{97.53} & \underline{94.76} & \underline{97.16} & \textbf{0.111} \\
    & PaddleOCR-VL-1.6~\cite{zhang2026paddleocr} & 0.9B & 96.33 & 0.033 & \underline{97.49} & \underline{94.76} & 97.11 & 0.127 \\
    & MinerU2.5-Pro~\cite{wang2026mineru2} & 1.2B & 95.75 & 0.036 & 97.45 & 93.42 & 95.92 & 0.120 \\
    & GLM-OCR~\cite{glm_ocr} & 0.9B & 95.22 & 0.044 & 97.18 & 92.83 & 95.39 & 0.133 \\
    & PaddleOCR-VL-1.5~\cite{paddleocr_vl15} & 0.9B & 94.87 & 0.038 & 96.69 & 91.67 & 94.37 & 0.130 \\
    & HunyuanOCR-1.5~\cite{li2026hunyuanocr} & 1B & 94.74 & 0.033 & 97.49 & 94.76 & 97.11 & 0.127 \\
    & PaddleOCR-VL~\cite{cui2025paddleocrvl} & 0.9B & 94.11 & 0.040 & 95.70 & 90.65 & 93.74 & 0.135 \\
    & Youtu-Parsing~\cite{YoutuParsing} & 2.5B & 93.68 & 0.044 & 93.45 & 92.02 & 95.00 & 0.116 \\
    & Logics-Parsing-v2~\cite{LogicsParsing} & 4B & 93.27 & 0.041 & 95.47 & 88.42 & 91.98 & 0.137 \\
    & FireRed-OCR~\cite{wu2026fireredocrtechnicalreport} & 2B & 93.20 & 0.037 & 95.27 & 88.04 & 91.06 & 0.131 \\
    & MinerU2.5 ~\cite{mineru25} & 1.2B & 92.98 & 0.045 & 95.59 & 87.88 & 91.47 & 0.130 \\
    & OpenDoc-0.1B~\cite{du2025unirec} & 0.1B & 90.64 & 0.049 & 92.93 & 83.88 & 87.45 & 0.140 \\
    & dots.ocr~\cite{dots_ocr} & 3B & 90.50 & 0.048 & 89.12 & 87.18 & 90.58 & 0.138 \\
    & DeepSeek-OCR 2~\cite{wei2025deepseek} & 3B & 90.17 & 0.050 & 91.59 & 83.89 & 87.75 & 0.144 \\
    & HunyuanOCR~\cite{hunyuanvisionteam2025hunyuanocrtechnicalreport} & 1B & 89.87 & 0.089 & 87.44 & 91.01 & 93.23 & 0.171 \\  
    & Dolphin-v2~\cite{feng2025dolphin} & 3B & 89.34 & 0.069 & 90.53 & 84.40 & 87.44 & 0.150 \\
    & OCRVerse~\cite{zhong2026ocrverseholisticocrendtoend} & 4B & 88.44 & 0.063 & 89.14 & 82.44 & 86.27 & 0.163 \\
    & MonkeyOCR-pro-3B~\cite{monkeyocr} & 3B & 88.43 & 0.074 & 88.33 & 84.35 & 88.62 & 0.189 \\
    \midrule
    \multirow{6}{*}{\makecell{\textbf{General}\\\textbf{VLMs}}}
    & Ovis2.6-30B-A3B~\cite{lu2024ovisstructuralembeddingalignment,lu2025ovis25technicalreport} & 30B & 93.62 & 0.035 & 94.93 & 89.44 & 92.40 & 0.135 \\
    & Gemini 3 Pro & -- & 92.85 & 0.064 & 95.83 & 89.15 & 92.96 & 0.165 \\
    & Gemini 3 Flash & -- & 92.58 & 0.066 & 95.03 & 89.29 & 93.51 & 0.173 \\
    & Qwen3-VL-235B~\cite{yang2025qwen3} & 235B & 89.78 & 0.063 & 92.53 & 83.07 & 86.75 & 0.166 \\
    & GPT-5.2 & -- & 86.52 & 0.114 & 88.00 & 82.95 & 87.93 & 0.193 \\
    & InternVL3.5-241B~\cite{wang2025internvl35advancingopensourcemultimodal} & 241B & 83.61 & 0.130 & 89.52 & 74.35 & 79.78 & 0.215 \\
    \bottomrule
    \end{tabular}%
  }
  \label{tab:omni_result}
  \vspace{-4pt}
\end{table*}

\textbf{OmniDocBench v1.6} is a representative benchmark for page-level digital document parsing, consisting of 1,651 PDF pages across 10 document categories, 5 layout types, and 5 languages. Compared with v1.5, v1.6 introduces a refined element matching strategy for formula evaluation and a challenging subset containing structurally complex pages, enabling more discriminative evaluation of advanced parsing models. The benchmark evaluates four key aspects of document parsing: text recognition using normalized edit distance, formula recognition using the Character Detection Metric (CDM)~\cite{wang2025image}, table reconstruction using Tree Edit Distance Similarity (TEDS) and TEDS-S~\cite{zhong2020image}, and reading order recovery using text block sequence edit distance. The overall score is computed as the average of text recognition, formula CDM, and table TEDS.

As shown in Table~\ref{tab:omni_result}, TeleOCR outperforms existing pipeline-based methods, including PaddleOCR-VL-1.6, MinerU2.5-Pro, and GLM-OCR, as well as the end-to-end approach OvisOCR2 on OmniDocBench v1.6. Specifically, TeleOCR achieves the best performance in text recognition, table reconstruction, and reading order recovery, obtaining the lowest normalized edit distance for text and reading order evaluation, and the highest TEDS and TEDS-S scores for table parsing. For formula recognition, TeleOCR also achieves competitive results.Further analysis shows that most formula recognition errors are caused by inconsistencies between the first-stage layout parsing results and the granularity of official OmniDocBench annotations. This observation suggests that more fine-grained layout modeling is required for further improvement. For table parsing, the proposed content-structure decoupled learning strategy enhances the modeling of complex table structures and improves reconstruction stability. Overall, TeleOCR demonstrates strong document parsing capability and robustness across diverse document understanding tasks.

\textbf{PureDocBench-Clean} is a comprehensive benchmark for evaluating document parsing under diverse acquisition conditions. It generates document images by rendering HTML source files and directly derives annotations from the source files, avoiding manual annotation errors while ensuring annotation consistency. The benchmark contains 1,475 pages from 10 domains and 66 subcategories, with three evaluation tracks: \textbf{Clean} for original rendered pages, \textbf{Digital} for degraded digital documents, and \textbf{Real} for document images captured from physical media or screens. As shown in Table~\ref{tab:puredoc}, TeleOCR achieves an overall score of 86.90 on the \textbf{Clean} track, surpassing the end-to-end method OvisOCR2 and demonstrating strong performance on high-quality digital documents.


\subsection{Camera-Captured Document Parsing}
\textbf{Wild-OmniDocBench} is a benchmark for evaluating the robustness of document parsing models under real-world capture conditions. Built upon OmniDocBench v1.5~\cite{ouyang2024omnidocbenchbenchmarkingdiversepdf}, it transforms digital documents into naturally captured images through a physical simulation pipeline involving document printing, deformation, and image acquisition under diverse illumination conditions. Unlike conventional benchmarks based on clean scanned documents or digital renderings, Wild-OmniDocBench introduces realistic degradations, including geometric distortions, illumination variations, screen-capture artifacts, and environmental noise. As shown in Table~\ref{tab:Wild_result}, TeleOCR achieves state-of-the-art performance on the real-capture track of Wild-OmniDocBench, outperforming existing end-to-end document parsing methods and demonstrating strong robustness in practical scenarios.

\textbf{PureDocBench-Degraded} is constructed from the electronic PDF documents in \textbf{PureDocBench-Clean} by simulating digital degradation processes and real-world acquisition conditions. With complex geometric distortions, diverse degradation patterns, and highly structured layouts, this benchmark provides a challenging testbed for evaluating the robustness of document parsing models. By introducing deformation-aware perception and modeling mechanisms, TeleOCR achieves state-of-the-art performance on the Degraded track in Table~\ref{tab:puredoc}, outperforming existing end-to-end document parsing methods.

\begin{table*}[h]
  \centering
\caption{Performance comparison of document parsing methods on Wild OmniDocBench~v1.5 Full for camera-captured document parsing across text, formula, table, and reading order extraction.}
  \resizebox{1\textwidth}{!}{
    \begin{tabular}{c|l l|c|c c c c c}
    \toprule
    \textbf{Model Type} & \textbf{Methods} & \textbf{Param} & \textbf{Overall}$\uparrow$ & \textbf{Text\textsuperscript{Edit}}$\downarrow$ & \textbf{Formula\textsuperscript{CDM}}$\uparrow$ & \textbf{Table\textsuperscript{TEDS}}$\uparrow$ & \textbf{Table\textsuperscript{TEDS-S}}$\uparrow$ & \textbf{Read Order\textsuperscript{Edit}}$\downarrow$ \\
    \midrule
    \multirow{5}{*}{\makecell{\textbf{Decoupled}\\\textbf{VLMs}}} 
    & \cellcolor{gray!20}\textbf{TeleOCR} & \cellcolor{gray!20}1.2B & \cellcolor{gray!20}\textbf{88.53} & \cellcolor{gray!20}\textbf{0.1173} & \cellcolor{gray!20}88.26 & \cellcolor{gray!20}\textbf{89.05} & \cellcolor{gray!20}\textbf{92.14} & \cellcolor{gray!20}\textbf{0.2011} \\
    & PaddleOCR-VL-1.6~\cite{zhang2026paddleocr} & 0.9B & 87.36 & 0.1369 & 88.42 & \underline{85.76} & \underline{90.14} & 0.2057 \\
    & MinerU2.5-Pro~\cite{wang2026mineru2} & 1.2B & 87.33 & 0.1362 & \underline{90.15} & 85.46 & 90.12 & \underline{0.2013} \\
    & GLM-OCR~\cite{glm_ocr} & 0.9B & 85.08 & 0.1514 & 89.09 & 81.31 & 85.90 & 0.2228 \\
    & PaddleOCR-VL-1.5~\cite{paddleocr_vl15} & 0.9B & 84.64 & 0.1461 & 86.72 & 81.80 & 86.52 & 0.2138 \\
    \midrule
    \multirow{4}{*}{\makecell{\textbf{End-to-End}\\\textbf{VLMs}}}
    & OvisOCR2~\cite{lu2026ovisocr2} & 0.8B & \underline{87.91} & 0.129 & \textbf{90.37} & 85.13 & 89.11 & 0.2021 \\
    & dots.ocr~\cite{dots_ocr} & 3B & 81.84 & 0.1483 & 85.0 & 75.32 & 80.20 & 0.2200 \\
    & HunyuanOCR-1.5~\cite{li2026hunyuanocr} & 1B & 77.62 & 0.1979 & 85.12 & 67.54 & 70.67 & 0.2750 \\
    & Logics-Parsing-v2~\cite{LogicsParsing} & 4B & 77.10 & 0.4029 & 91.4 & 80.19 & 87.16 & 0.2355 \\
    \bottomrule
    \end{tabular}%
  }
  \label{tab:Wild_result}
  \vspace{-4pt}
\end{table*}


\begin{table*}[t]
\centering
\caption{
Comparison with existing document parsing models under clean and degraded scenarios.
$\uparrow$ indicates higher is better, while $\downarrow$ indicates lower is better.
}
\resizebox{\linewidth}{!}{
\begin{tabular}{lcccccccccccc}
\toprule
\multirow{2}{*}{Model}
& \multicolumn{4}{c}{Clean}
& \multicolumn{4}{c}{Digital Degraded}
& \multicolumn{4}{c}{Real Degraded}
\\
\cmidrule(lr){2-5}
\cmidrule(lr){6-9}
\cmidrule(lr){10-13}
&
Overall$\uparrow$
& Text$\downarrow$
& Formula$\uparrow$
& Table$\uparrow$
&
Overall$\uparrow$
& Text$\downarrow$
& Formula$\uparrow$
& Table$\uparrow$
&
Overall$\uparrow$
& Text$\downarrow$
& Formula$\uparrow$
& Table$\uparrow$
\\
\midrule
\multicolumn{13}{c}{\textit{Decoupled VLM}}
\\
\midrule
    
\cellcolor{gray!20}\textbf{TeleOCR}
& \cellcolor{gray!20}\textbf{86.90} & \cellcolor{gray!20}\textbf{0.111} & \cellcolor{gray!20}\textbf{81.01} & \cellcolor{gray!20}\textbf{91.09}
& \cellcolor{gray!20}\underline{77.47} & \cellcolor{gray!20}0.206 & \cellcolor{gray!20}\textbf{72.59} & \cellcolor{gray!20}80.45
& \cellcolor{gray!20}\textbf{70.85} & \cellcolor{gray!20}\underline{0.302} & \cellcolor{gray!20}\textbf{65.11} & \cellcolor{gray!20}\textbf{77.66}
\\

DotsMOCR~\cite{zheng2026multimodal}
&76.27&0.151&66.23&77.65
&73.16&\underline{0.198}&64.32&74.95
&61.73&0.312&54.39&61.97
\\

MinerU2.5-Pro~\cite{wang2026mineru2}
&75.87&0.222&65.14&84.68
&71.77&0.272&61.79&80.73
&62.56&0.375&52.70&72.47
\\

YouTu-Parsing~\cite{YoutuParsing}
&75.02&0.230&67.34&80.74
&69.66&0.270&61.44&74.49
&60.29&0.360&52.20&64.69
\\

PaddleOCR-VL-1.5~\cite{paddleocr_vl15}
&73.01&0.266&63.53&82.12
&66.73&0.339&58.03&76.07
&60.50&0.398&54.00&67.33
\\

GLM-OCR~\cite{glm_ocr}
&68.65&0.314&57.89&79.44
&63.06&0.383&53.23&74.21
&58.31&0.433&50.34&67.83
\\

Dolphin-v2~\cite{feng2025dolphin}
&65.90&0.342&59.80&72.12
&60.24&0.393&52.20&67.86
&44.92&0.553&39.98&50.04
\\

MonkeyOCR-pro-3B~\cite{monkeyocr}
&62.23&0.346&48.46&72.83
&57.40&0.397&45.57&66.32
&46.49&0.511&38.18&52.43
\\

\midrule
\multicolumn{13}{c}{\textit{End-to-End VLM}}
\\
\midrule

OvisOCR2~\cite{lu2026ovisocr2}
&\underline{82.14}&\underline{0.149}&\underline{71.29}&\underline{90.12}
&\textbf{77.77}&\textbf{0.192}&\underline{67.87}&\textbf{84.71}
&66.61&0.316&57.64&\underline{73.79}
\\

FD-RL~\cite{zhong2026reading}
&78.38&0.193&68.21&86.22
&76.33&0.214&67.16&\underline{83.22}
&\underline{67.04}&\textbf{0.298}&58.82&72.08
\\

Logics-Parsing-v2~\cite{LogicsParsing}
&76.35&0.213&67.67&82.67
&73.85&0.248&67.33&79.02
&67.64&0.304&\underline{61.65}&71.64
\\

dots.ocr~\cite{dots_ocr}
&72.01&0.248&61.37&79.51
&65.95&0.307&56.67&71.86
&55.68&0.403&47.70&59.63
\\

Qianfan-OCR~\cite{dong2026qianfan}
&57.22&0.370&49.79&58.83
&50.85&0.438&44.41&51.96
&45.06&0.494&39.08&45.53
\\

\midrule
\multicolumn{13}{c}{\textit{General VLMs}}
\\
\midrule

Qwen3-VL-8B~\cite{yang2025qwen3}
&72.44&0.261&65.10&78.35
&72.03&0.266&64.88&77.82
&62.73&0.342&55.55&66.81
\\

Kimi K2.6
&72.32&0.303&66.93&80.30
&69.95&0.322&64.69&77.31
&68.02&0.335&62.44&75.14
\\

Gemini-3.1-Pro
&70.04&0.306&65.63&75.08
&69.28&0.322&65.81&74.24
&71.98&0.300&68.62&77.26
\\

Qwen3.5-397B-A17B~\cite{team2026qwen3}
&69.12&0.233&65.26&65.40
&68.34&0.244&63.91&65.53
&62.70&0.287&60.70&56.12
\\
\bottomrule
\end{tabular}
}
\label{tab:puredoc}
\end{table*}

\subsection{ Scientific Figure-to-Table conversion}
The Sci-ImageMiner Challenge~\cite{ahmed2026icdar} focuses on scientific image understanding in real-world research papers, with an emphasis on quantitative analysis of scientific figures in the Atomic Layer Deposition and Etching (ALD/E) domain. The challenge aims to bridge the gap between visual content understanding and scientific data interpretation. Among its tasks, Scientific Figure-to-Table conversion is a key task that requires models to recover structured experimental data from scientific figures by transforming visual elements, including curves, axes, and legends, into machine-readable tables.

TeleOCR improves the joint modeling of structural and semantic information in scientific figures through a content-structure decoupled learning strategy. It achieves the best TEDS performance on the Scientific Figure-to-Table task, outperforming the second-best method by over 2 percentage points. These results demonstrate that TeleOCR extends beyond general document parsing and exhibits strong generalization capability in specialized scientific domains.

\begin{table}[htb]
\centering
\scriptsize
\caption{Data Extraction performance comparison among the top-5 teams and the best baseline in the ICDAR 2026 Sci-ImageMiner Challenge.}
\resizebox{0.7\linewidth}{!}{
    \begin{tabular}{l l c c c}
    \toprule
    \textbf{\#} & \textbf{Team} & \textbf{RMS} & \textbf{TEDS} & \textbf{Weighted} \\
    \midrule
    \cellcolor{gray!20}1 & \cellcolor{gray!20}TeleOCR & \cellcolor{gray!20}\textbf{17.23} & \cellcolor{gray!20}\textbf{66.39} & \cellcolor{gray!20}\textbf{41.81} \\
    2 & VLMinators & 17.29 & 64.31 & 40.80 \\
    3 & Ricoh\_SRCB & 16.23 & 61.12 & 38.67 \\ 
    4 & Vassilis Sioros & 14.94 & 55.20 & 35.07 \\
    5 & DocMiner & 12.67 & 53.72 & 33.19 \\
    6 & Qwen3 VL 8b~\cite{yang2025qwen3} & 14.08 & 57.86 & 35.97 \\
    \bottomrule
    \end{tabular}
}
\label{tab:leaderboard_data_extraction}
\end{table}

\section{Conclusion}
\label{sec:conclusion}
This paper presents TeleOCR: Navigating Document Parsing Across Digital and Camera-Captured Documents, a unified document parsing framework designed to achieve robust understanding of both digital and camera-captured documents. To overcome the limitations of existing OCR systems in complex layouts, structured content parsing, and real-world acquisition scenarios, TeleOCR introduces comprehensive improvements from three aspects: data construction, training strategies, and model capabilities.

First, we establish a large-scale document data engineering pipeline covering diverse tasks, including text recognition, layout analysis, table parsing, formula recognition, code recognition, and scientific figure understanding. Through data cleaning, synthetic data generation, and model-assisted verification, the quality of training data is enhanced, improving the model's generalization ability across diverse document scenarios. Second, we propose a structure-aware progressive training strategy, where content-structure decoupled learning enhances the model's capability to represent complex document structures, including table layouts, formula syntax, and intricate page designs. Meanwhile, deformation-aware learning and adaptive sampling mechanisms are introduced to enable effective handling of perspective distortions, irregular layouts, and low-quality captured documents. Furthermore, multi-model consistency verification and self-evaluation mechanisms are employed to automatically filter high-quality training samples, further improving data reliability.

TeleOCR supports unified parsing of diverse document elements, including text, tables, formulas, code blocks, seals, and scientific figures, enabling end-to-end transformation from document images to structured information. Extensive evaluations on multiple public benchmarks and real-world scenarios demonstrate the strong performance of TeleOCR, validating its effectiveness and generalization capability for both digital and camera-captured document parsing.

\bibliographystyle{plainnat}
\bibliography{cite}
\newpage
\beginappendix

\section{Prompt Design and Task Examples}
This section presents the prompt formats, output specifications, and representative examples of the tasks supported by TeleOCR. All tasks follow a unified prompt interface, where each input contains only an \texttt{<image>} token followed by a textual task instruction, without requiring additional few-shot examples or structured metadata. TeleOCR supports 8 document parsing tasks, with their corresponding instructions and output formats summarized below:
\begin{itemize}
\item \textbf{Digital Layout Detection} (\S\ref{app:Digital_Layout_Detection}) — Detects content regions in digital documents and outputs structured detection results containing bounding boxes, category labels, and rotation directions.
\item \textbf{Camera-captured Layout Segmentation} (\S\ref{app:Camera_captured_Layout_Segmentation}) — Localizes content regions in camera-captured documents and outputs polygon-based region boundaries, category labels, and rotation directions.
\item \textbf{Text Recognition} (\S\ref{app:text_recognition}) — Transcribes cropped text regions into corresponding text sequences.
\item \textbf{Formula Recognition} (\S\ref{app:formula_recognition}) — Converts cropped formula regions into LaTeX representations.
\item \textbf{Table Recognition} (\S\ref{app:table_recognition}) — Converts cropped tables into structured Token sequences based on OTSL, including cell contents, which are further parsed into HTML representations.
\item \textbf{Code Block Recognition} (\S\ref{app:Code_Block_Recognition}) — Converts cropped code regions into Markdown format and simultaneously predicts the corresponding programming language type.
\item \textbf{Scientific Figure Analysis} (\S\ref{app:Scientific_Figure_Analysis}) — Converts cropped scientific figures into structured table Token sequences represented by OTSL.
\item \textbf{Seal Recognition} (\S\ref{app:Seal_Recognition}) — Transcribes cropped seal regions into text sequences.
\end{itemize}

\subsection{Digital Layout Detection}
\label{app:Digital_Layout_Detection}
TeleOCR retains the layout parsing capability for digital documents, enabling precise localization of structured regions within a page. This task outputs the rectangular bounding box, semantic category, and text orientation for each detected region. The model takes a downsampled page image as input and generates a structured layout representation composed of multiple region descriptions.

\paragraph{Prompt.}
\begin{verbatim}
<image>\nAnalyze the image layout.
\end{verbatim}

\paragraph{Output Format.}

The model outputs a sequence of region descriptions separated by newline characters, where each region follows the unified format:

\begin{verbatim}
<box:x1 y1 x2 y2><label:category><rotate_dir>
\end{verbatim}

This output format adapts the representation introduced in MinerU by compacting region descriptions to reduce token consumption, while providing a unified interface for both digital and camera-captured document layout parsing tasks. Specifically, \texttt{x1 y1 x2 y2} denote the normalized coordinates of the rectangular bounding box, mapped to a $[0,999]$ grid space; \texttt{category} represents the semantic category label of the region; and \texttt{<rotate\_dir>} indicates the text orientation.

\subsection{Camera-captured Layout Segmentation}
\label{app:Camera_captured_Layout_Segmentation}
TeleOCR extends layout parsing to camera-captured document scenarios by integrating both global point-level and local region-level deformation-aware modeling capabilities. This task aims to localize structured regions in camera-captured documents and output the polygonal boundary points, semantic categories, and text orientations of each region.

\paragraph{Prompt.}
\begin{verbatim}
<image>\nMulti-point Layout Segmentation Analysis.
\end{verbatim}

\paragraph{Output Format.}

The model outputs a sequence of region descriptions separated by newline characters, where each region follows the unified format:

\begin{verbatim}
<box:x1 y1 x2 y2 x3 y3 ... ><label:category><rotate_dir>
\end{verbatim}

This representation maintains consistency with the digital layout detection task while extending rectangular bounding boxes to variable-length polygonal point sets. Specifically, the number of sampled coordinates is adaptively determined according to the local deformation complexity of each document region. Simple regions are represented with fewer points, whereas regions with complex non-rigid deformations are described using additional sampling points to capture finer geometric structures.

\subsection{Text Recognition}
\label{app:text_recognition}

The text recognition task aims to convert cropped text regions into corresponding text sequences. The input consists of cropped regions from both digital document layouts and camera-captured documents. For camera-captured documents, only the segmented foreground text regions are retained, while non-text areas are masked with black pixels to reduce interference from irrelevant visual content.

\paragraph{Prompt.}
\begin{verbatim} <image>\nPlease output the text content from the image.
\end{verbatim}

\paragraph{Output Format.}

The model outputs a plain-text sequence corresponding to the input text region while preserving structural information, including inline formulas, subscripts, superscripts, and special symbols.

\subsection{Formula Recognition}
\label{app:formula_recognition}

The formula recognition task aims to convert cropped formula regions into \LaTeX{} representations.

\paragraph{Prompt.}
\begin{verbatim}
<image>\nPlease write out the expression of the formula in the image using LaTeX format.
\end{verbatim}

\paragraph{Output Format.}

The model outputs a \LaTeX{} mathematical string containing standard commands and environments (e.g., \verb|\frac|, \verb|\mathrm|, and \verb|\quad|), which can be directly compiled. When equation numbers are present in the input image, the model preserves the corresponding numbering information using \verb|\tag{...}|.

\subsection{Table Recognition}
\label{app:table_recognition}

The table recognition task aims to convert cropped table regions into structured token sequences based on OTSL (Optimized Table Structure Language). Cell contents are transcribed as text, where inline formulas are represented using single dollar notation (\verb|$...$|). The generated OTSL sequence is further converted into an HTML representation for visualization and downstream applications.

\paragraph{Prompt.}

\begin{verbatim} <image>\nThis is the image of a table. Please output the table in OTSL format.
\end{verbatim}

\paragraph{Output Format.}

The model outputs a flattened token sequence representing the table structure, organized in row-major order. The OTSL representation provides a compact and unambiguous description of both regular grid tables and tables with complex cell structures. After generation, the OTSL sequence is automatically converted into HTML for table rendering and downstream system integration.

\begin{figure*}[!h]
\centering
\includegraphics[width=1\textwidth, trim=0mm 20mm 200mm 0mm, clip]{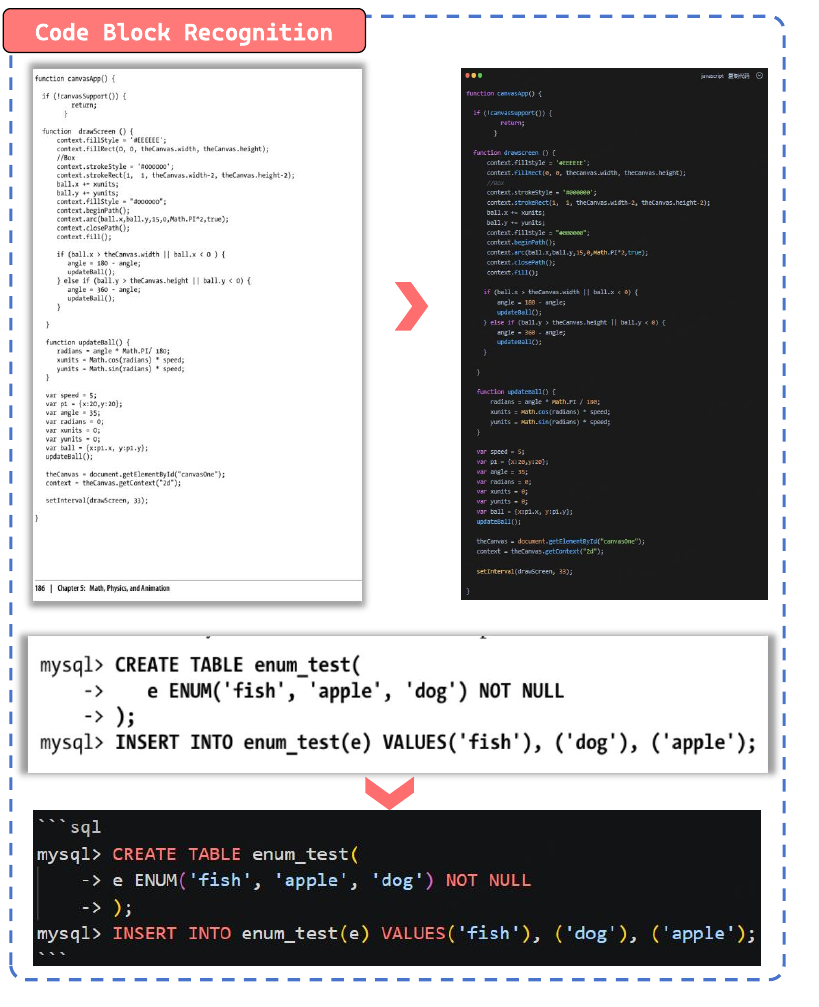} 
\caption{Case study of Code Block Recognition. TeleOCR achieves exact source code reconstruction and programming language identification.}
\label{fig:code_example}
\end{figure*}

\subsection{Code Block Recognition}
\label{app:Code_Block_Recognition}
The code block recognition task aims to recover source code from input code screenshots. It requires the model to preserve the original indentation, syntax structure, and formatting while identifying the programming language of the code block.

\paragraph{Prompt.}

\begin{verbatim} 
<image>\nThe image contains a code snippet, please output the parsing result.
\end{verbatim}

\paragraph{Output Format.}

The model outputs the recovered code in the Markdown code block format, where the first line specifies the programming language, followed by the corresponding source code:

\begin{verbatim}
```language
code
```
\end{verbatim}
\paragraph{Example.}
Given an example image from OmniDocBench v1.6, the corresponding model output is shown in the figure~\ref{fig:code_example}.

\begin{figure*}[t]
\centering
\includegraphics[width=1\textwidth, trim=0mm 0mm 80mm 0mm, clip]{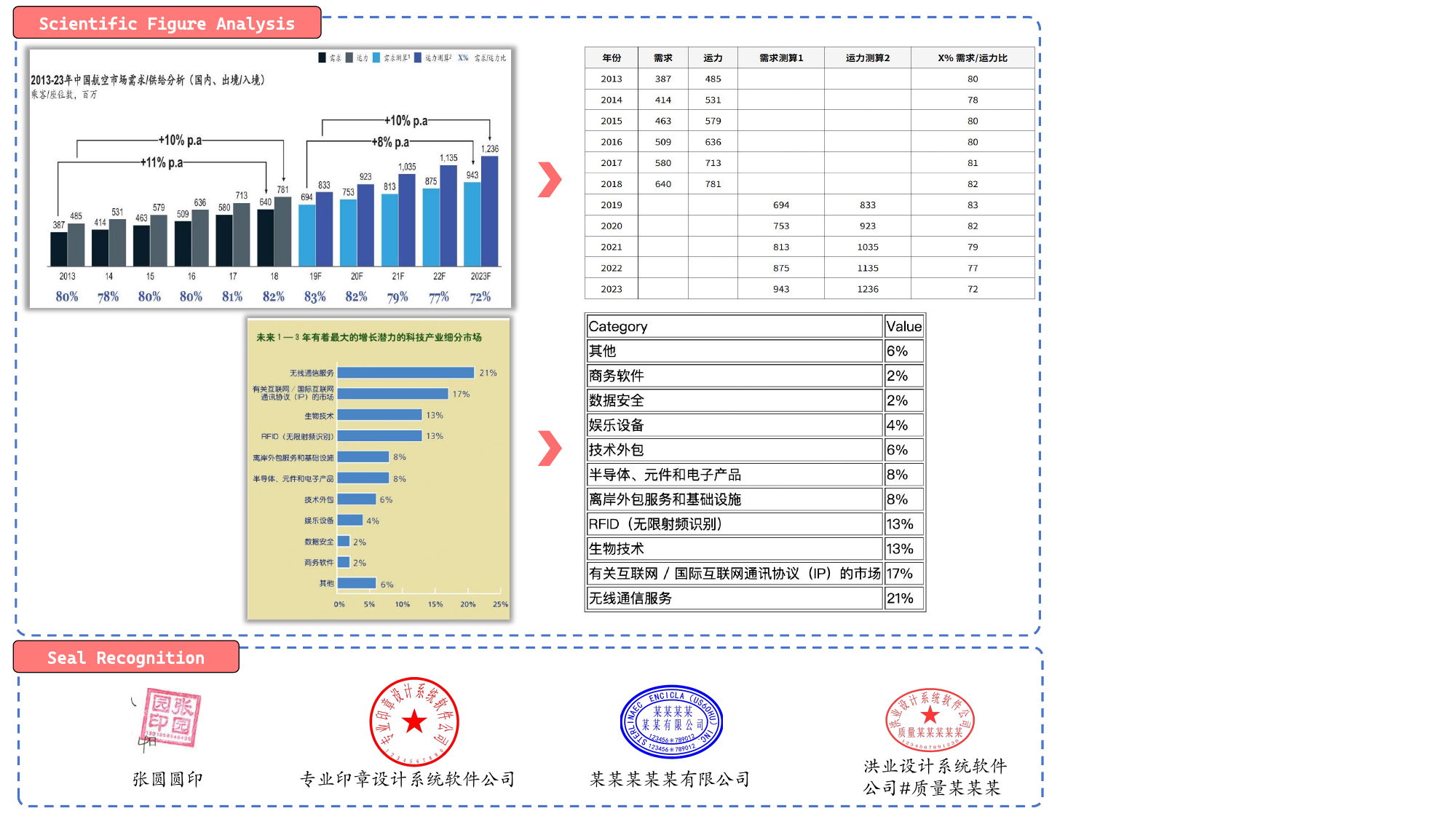} 
\caption{Case studies of Scientific Figure-to-Table and Seal Recognition. The scientific figures are collected from OmniDocBench v1.6, while the seals are obtained from anonymized Internet data.}
\label{fig:seal_figure}
\end{figure*}

\subsection{Scientific Figure Analysis}
\label{app:Scientific_Figure_Analysis}
The scientific figure analysis task aims to recover structured tabular data from scientific figures. TeleOCR supports scientific figure parsing across 5 major categories and 19 fine-grained classes. The five categories include univariate distribution, multivariate comparison, matrix-based, spatial localization, and structural flow figures. The fine-grained classes cover histograms, pie charts, donut charts, rose charts, tree diagrams, funnel charts, grouped bar charts, kernel density plots, bar distribution plots, stacked bar charts, stacked line charts, multi-line charts, radar charts, box plots, heatmaps, directed adjacency tables, undirected adjacency tables, bubble charts, and Sankey diagrams.

\paragraph{Prompt.}

\begin{verbatim} <image>\nThis is a scientific figure. Please extract the table implied by the figure.
\end{verbatim}

\paragraph{Output Format.}

The output format follows the table recognition task. The model generates a flattened OTSL token sequence representing the structured table, which is subsequently converted into a tabular representation. A representative example is shown in Figure~\ref{fig:seal_figure}.

\subsection{Seal Recognition}
\label{app:Seal_Recognition}
The seal recognition task aims to extract textual content from cropped seal regions. Due to their irregular shapes and interference from surrounding text, lines, and complex textures, seal regions pose challenges for accurate text recognition. TeleOCR focuses on extracting foreground seal text while suppressing irrelevant background information.

\paragraph{Prompt.}

\begin{verbatim} <image>\nSeal Recognition:
\end{verbatim}

\paragraph{Output Format.}

The model outputs only the textual content within the seal region, excluding irrelevant background text. A representative example is shown in Figure~\ref{fig:seal_figure}.

\section{Data Synthesis Details}

\begin{figure}[H]
\centering
\includegraphics[width=0.8\textwidth, trim=0mm 0mm 50mm 0mm, clip]{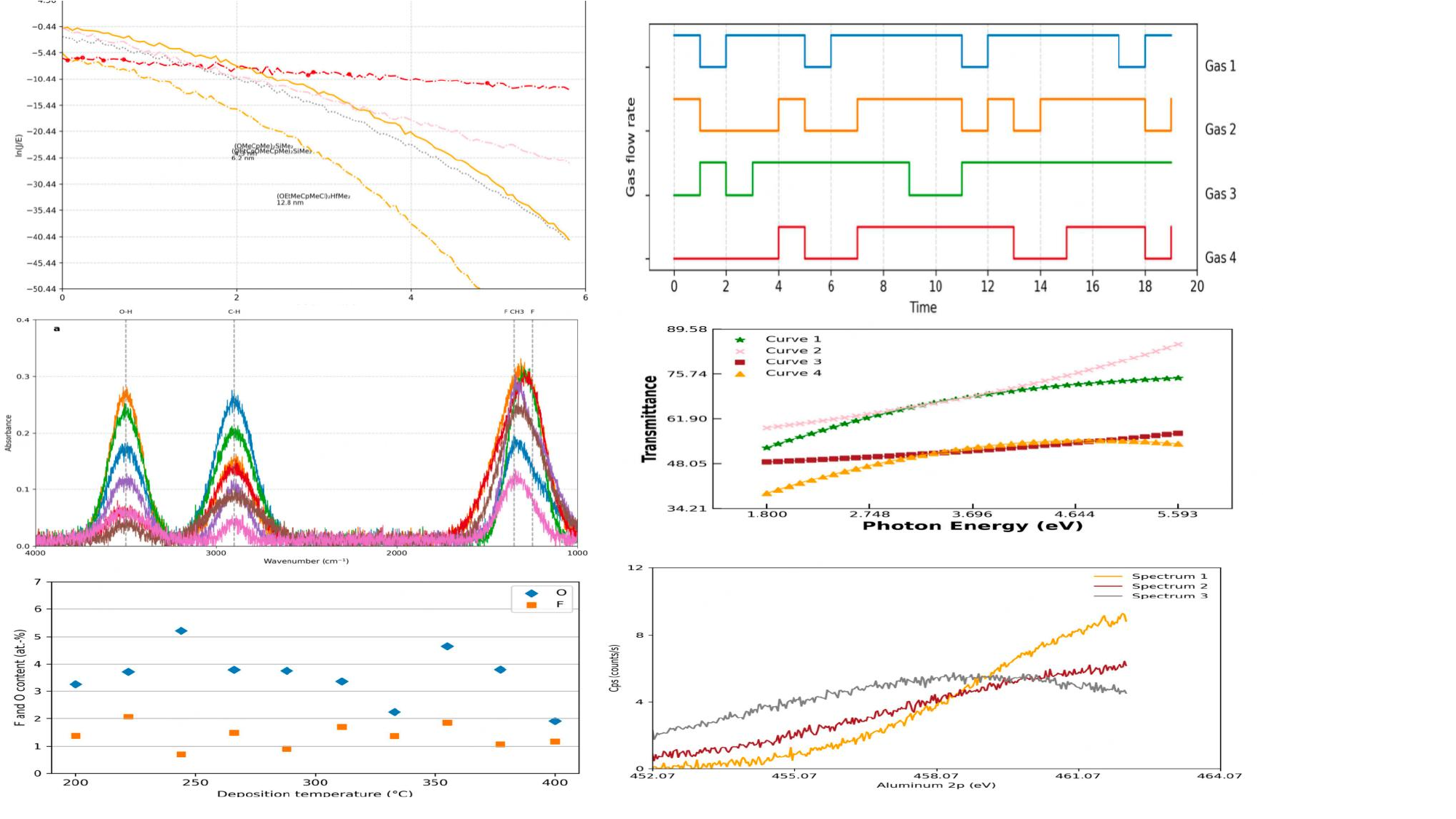} 
\includegraphics[width=0.8\textwidth, trim=0mm 0mm 150mm 0mm, clip]{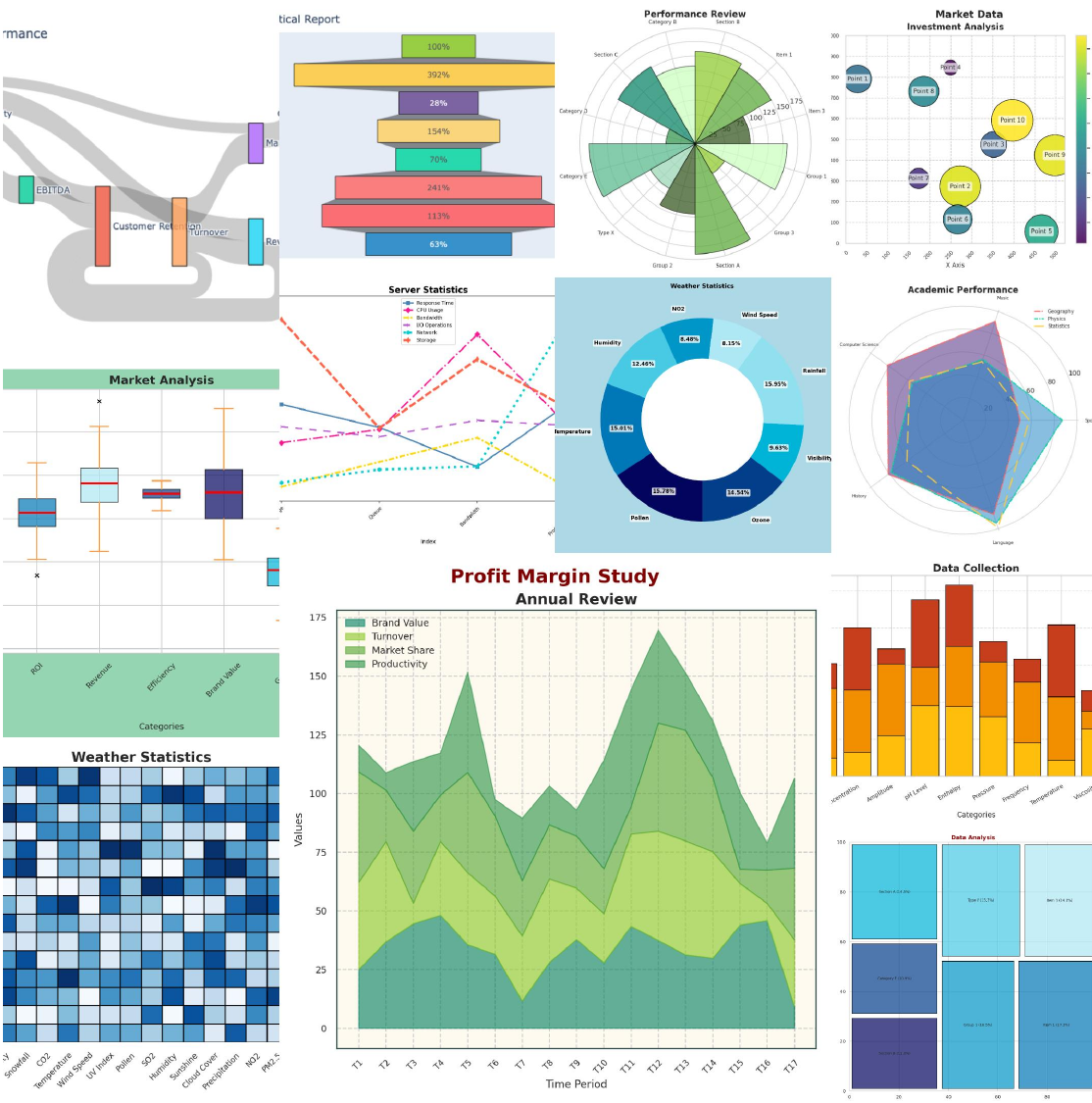} 
\caption{Case studies of synthetic data generation for Scientific Figure-to-Table.}
\label{fig:Scientific_figure}
\end{figure}

Due to the scarcity of large-scale, high-quality annotated scientific charts, we develop a synthetic data generation pipeline to enhance scientific chart understanding. Specifically, rendering engines such as Matplotlib are employed to automatically generate diverse scientific figures, including radar charts, heatmaps, line plots, scatter plots, and energy spectra. The generated samples are designed to mimic the visual characteristics and statistical distributions of real scientific publications. Additional examples are provided in Figure~\ref{fig:Scientific_figure}

The synthesis pipeline establishes a direct mapping between visual chart patterns and structured representations. Each synthetic chart is paired with its underlying numerical data, axis labels, legends, and corresponding data tables, providing end-to-end supervision from visual inputs to structured outputs. By controlling chart types, layouts, data distributions, and annotation styles, the generated dataset covers diverse scientific visualization scenarios.

To further improve structural understanding, we construct two complementary types of synthetic samples. The first type consists of full-content charts that preserve complete visual information, including values, labels, and legends, enabling accurate quantitative information extraction. The second type contains structure-only charts, where textual contents and numerical values are masked while preserving geometric layouts and structural relationships. These samples encourage the model to learn intrinsic chart structures, such as axis organization, spatial alignment, and hierarchical table topology.

This synthetic data generation strategy provides scalable supervision for scientific chart understanding and mitigates the limitation of insufficient real-world annotations. By jointly leveraging content-rich and structure-aware samples, the model achieves stronger generalization on diverse scientific figures from real research literature.

\section{Qualitative Comparison with SOTA Methods}
This section presents qualitative comparisons between TeleOCR and state-of-the-art methods across representative document scenarios, including native digital documents, digitally degraded documents, and real-world captured documents. The visual results evaluate the performance of layout analysis, text recognition, table parsing, and formula extraction.

\subsection{Layout Recognition}
\begin{figure}[H]
\centering
\includegraphics[width=1\textwidth, trim=0mm 0mm 70mm 0mm, clip]{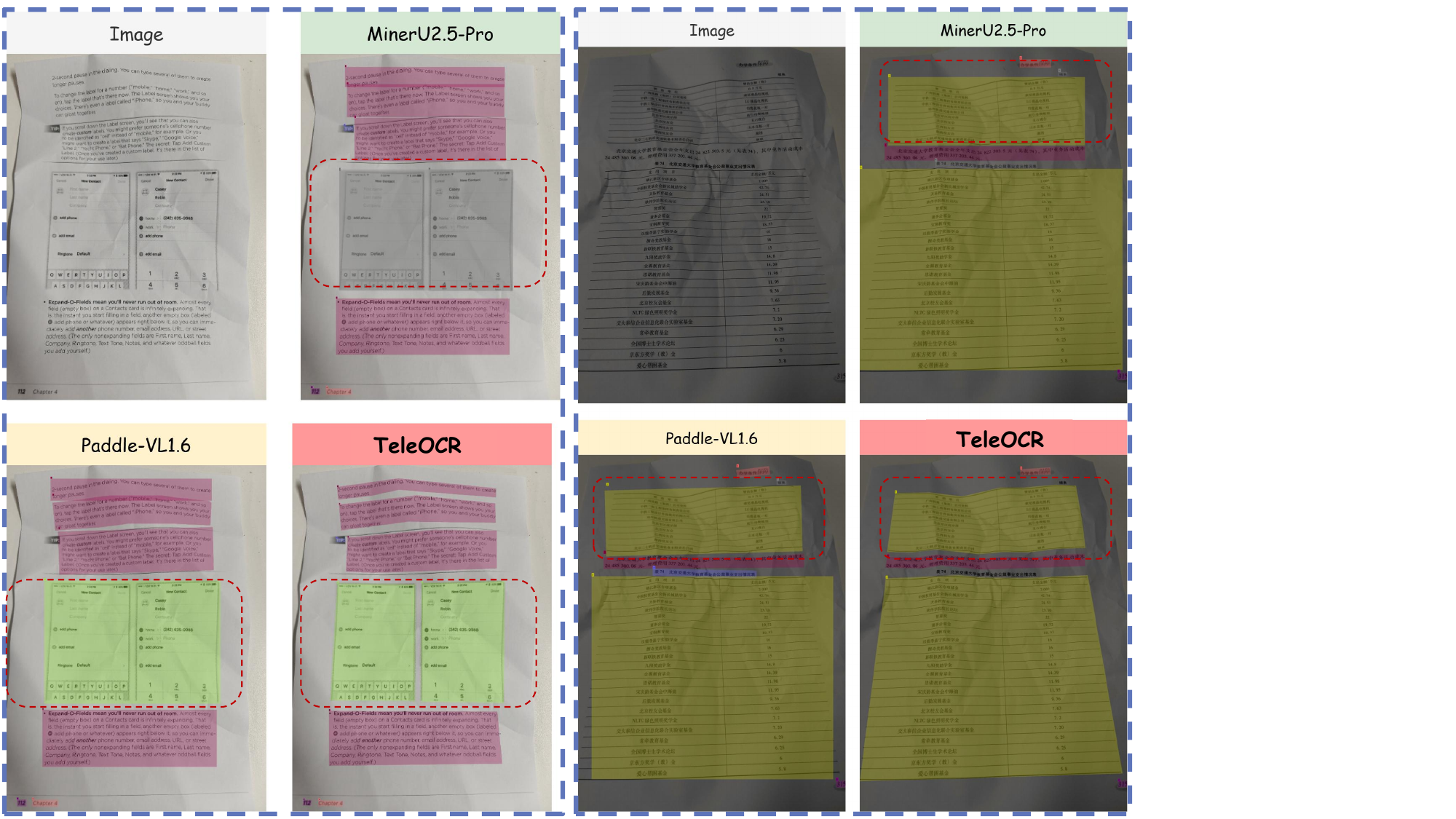} 
\caption{Qualitative comparison of layout recognition on captured structured documents. TeleOCR provides more reliable analysis of creased structured tables than other SOTA methods.}
\label{fig:layout_1}
\end{figure}

\begin{figure}[H]
\centering
\includegraphics[width=1\textwidth, trim=0mm 0mm 10mm 0mm, clip]{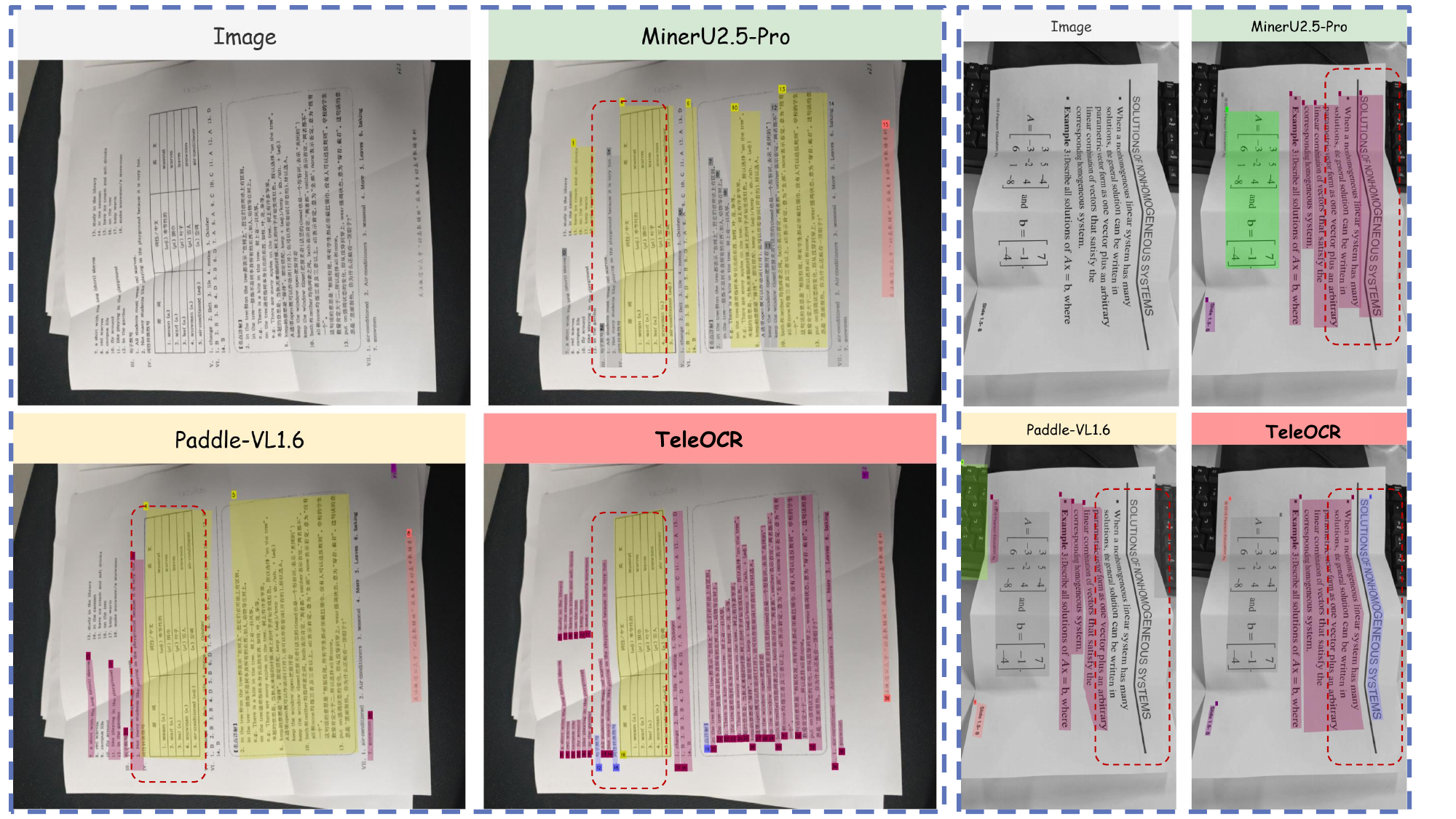} 
\caption{Qualitative comparison of layout recognition on rotated and creased documents. TeleOCR better covers creased regions and accurately recognizes small text areas.}
\label{fig:layout_2}
\end{figure}

\begin{figure}[H]
\centering
\includegraphics[width=1\textwidth, trim=0mm 0mm 30mm 0mm, clip]{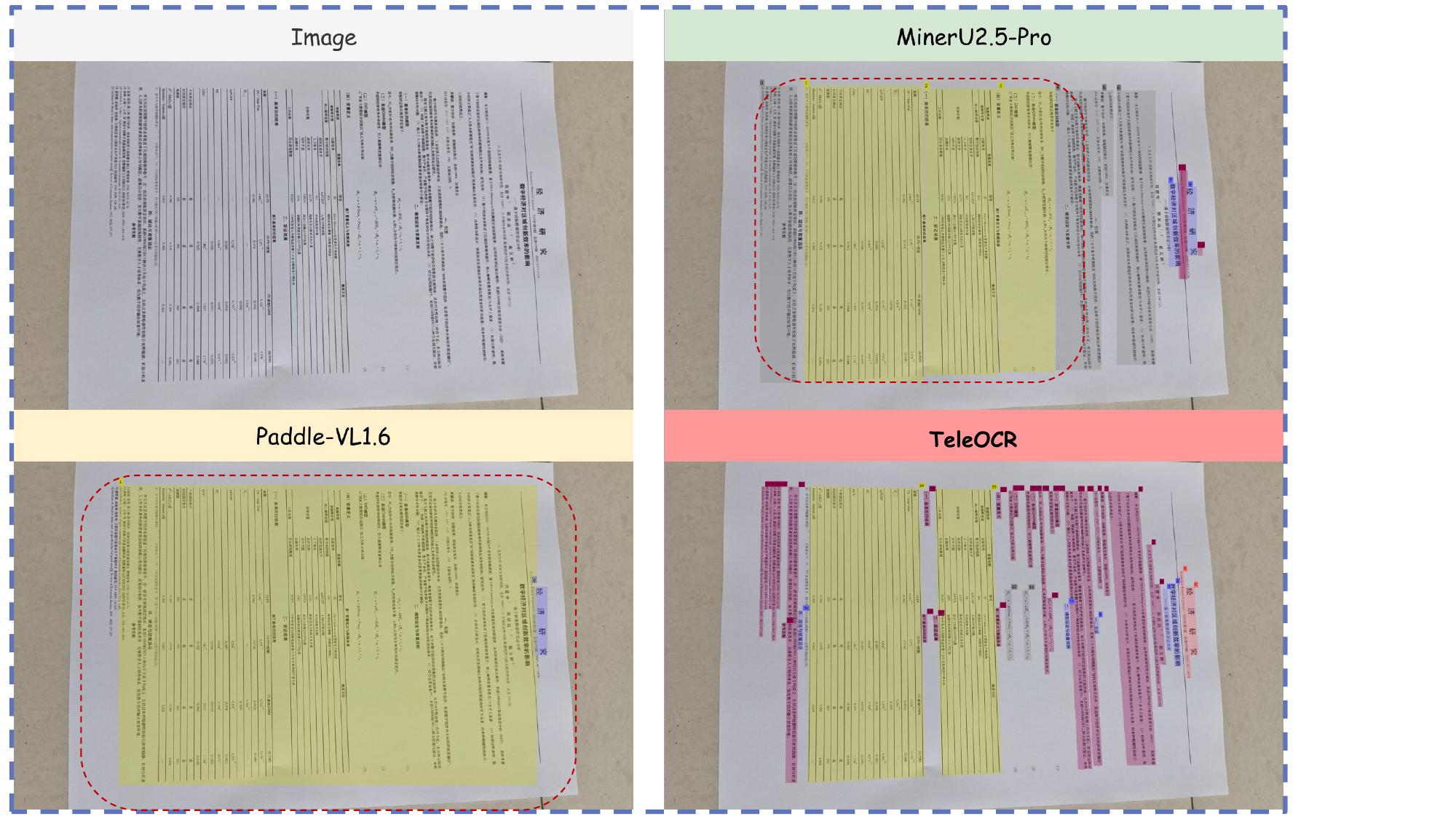} 
\caption{Qualitative comparison of layout recognition on complex captured documents. TeleOCR achieves more complete and fine-grained layouts for dense and complex document structures.}
\label{fig:layout_3}
\end{figure}

\begin{figure}[H]
\centering
\includegraphics[width=1\textwidth, trim=0mm 0mm 30mm 0mm, clip]{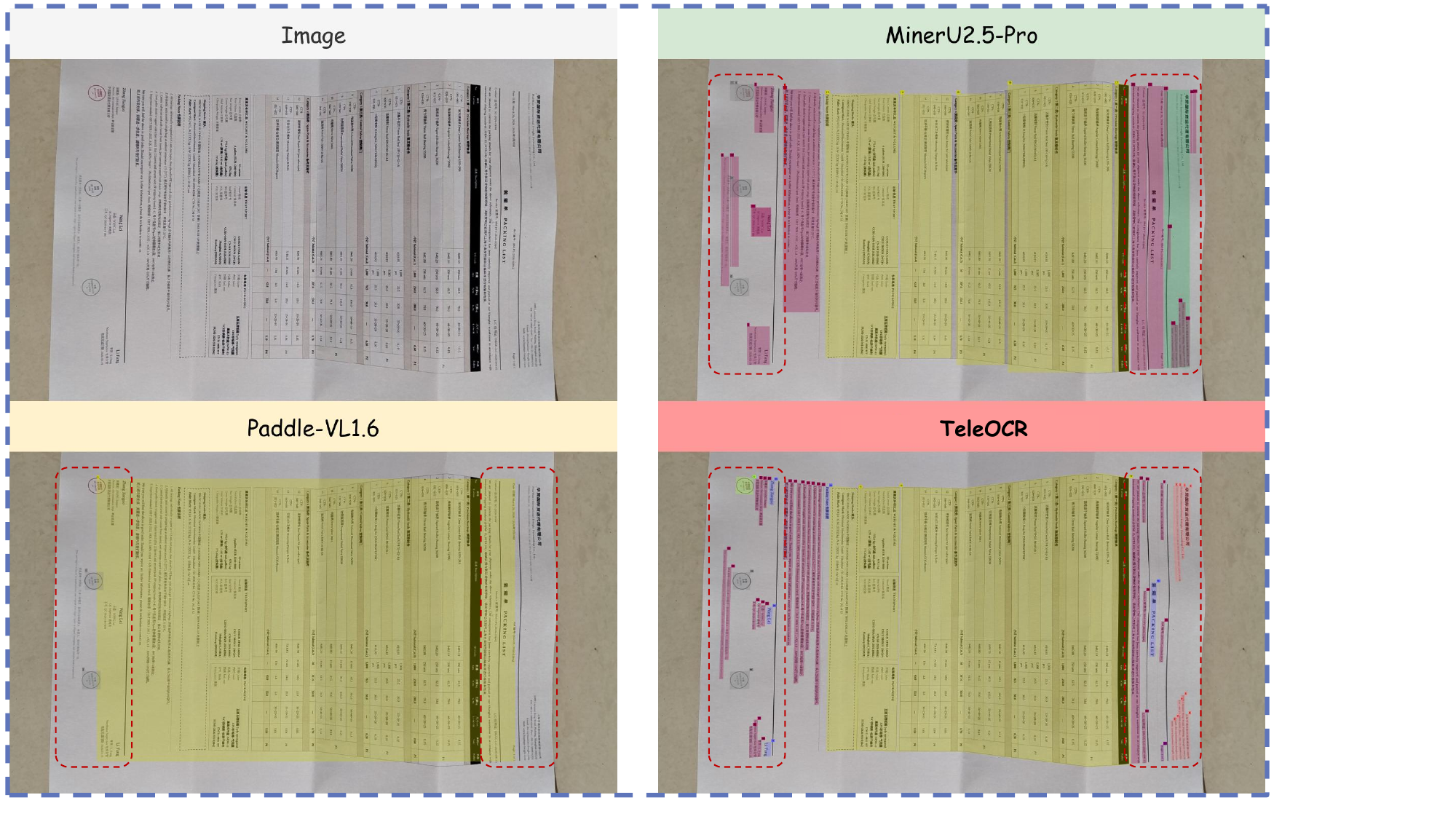} 
\caption{Qualitative comparison of layout recognition on severely distorted tables. TeleOCR enables structured layout parsing for highly distorted table documents.}
\label{fig:layout_4}
\end{figure}

TeleOCR adopts a decoupled parsing framework, where accurate layout recognition is essential for subsequent content understanding. We compare the layout prediction results of MinerU2.5 Pro~\cite{wang2026mineru2}, PaddleOCR-VL 1.6~\cite{zhang2026paddleocr}, and TeleOCR on real-world captured documents, as shown in Figures~\ref{fig:layout_1}, \ref{fig:layout_2}, \ref{fig:layout_3}, and \ref{fig:layout_4}

For documents with wrinkles, geometric distortions, and complex layouts, MinerU2.5 Pro~\cite{wang2026mineru2} and PaddleOCR-VL 1.6~\cite{zhang2026paddleocr} may produce missing regions or incorrect category predictions, limiting fine-grained layout understanding. In contrast, TeleOCR incorporates region-level and point-level deformation-aware learning with deformation augmentation to capture geometric variations in real-world documents, enabling more complete and accurate layout recognition, especially for dense text areas and complex table structures.


\subsection{Table Parsing}

We compare table parsing results under challenging camera-captured scenarios, including skew, perspective distortion, page curvature, blur, and dense table layouts. Overall, TeleOCR preserves table topology and cell relationships more reliably than competing methods, especially for distorted or fine-grained tables.



\begin{figure}[H]
\centering
\includegraphics[width=1\textwidth, trim=0mm 0mm 0mm 0mm, clip]{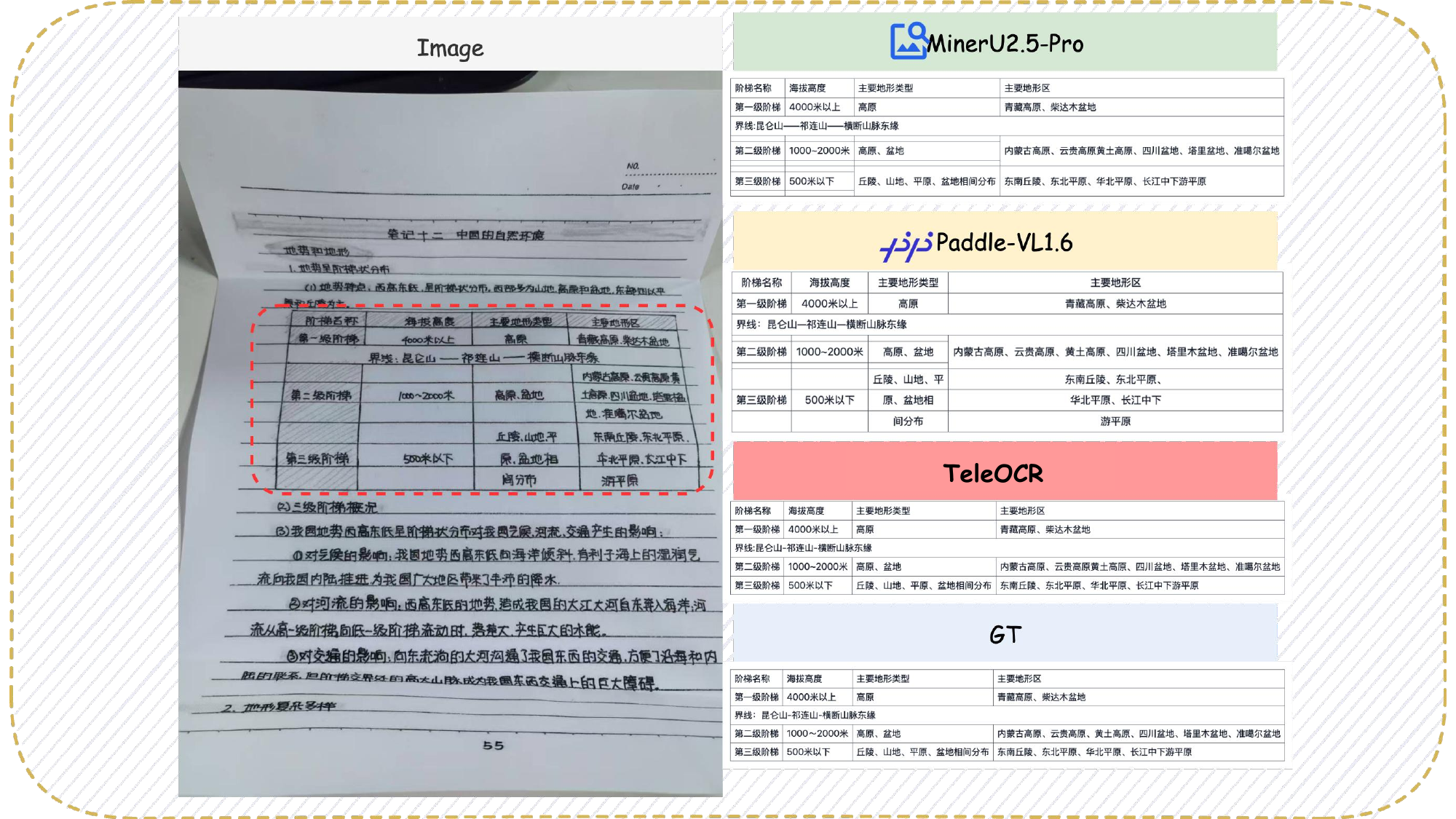}
\caption{Qualitative comparison on a distorted handwritten-note table. TeleOCR better preserves row-column alignment and cell correspondences under skew and blur.}
\label{fig:table_1}
\end{figure}

Figure \ref{fig:table_1} shows a handwritten note page with skew, perspective compression, and local blur. MinerU2.5-Pro and Paddle-VL-1.6 preserve part of the table content, but suffer from row-column misalignment and content merging. TeleOCR better recovers the four-column structure, showing stronger robustness to camera-captured distortions.

\begin{figure}[H]
\centering
\includegraphics[width=1\textwidth, trim=0mm 0mm 0mm 0mm, clip]{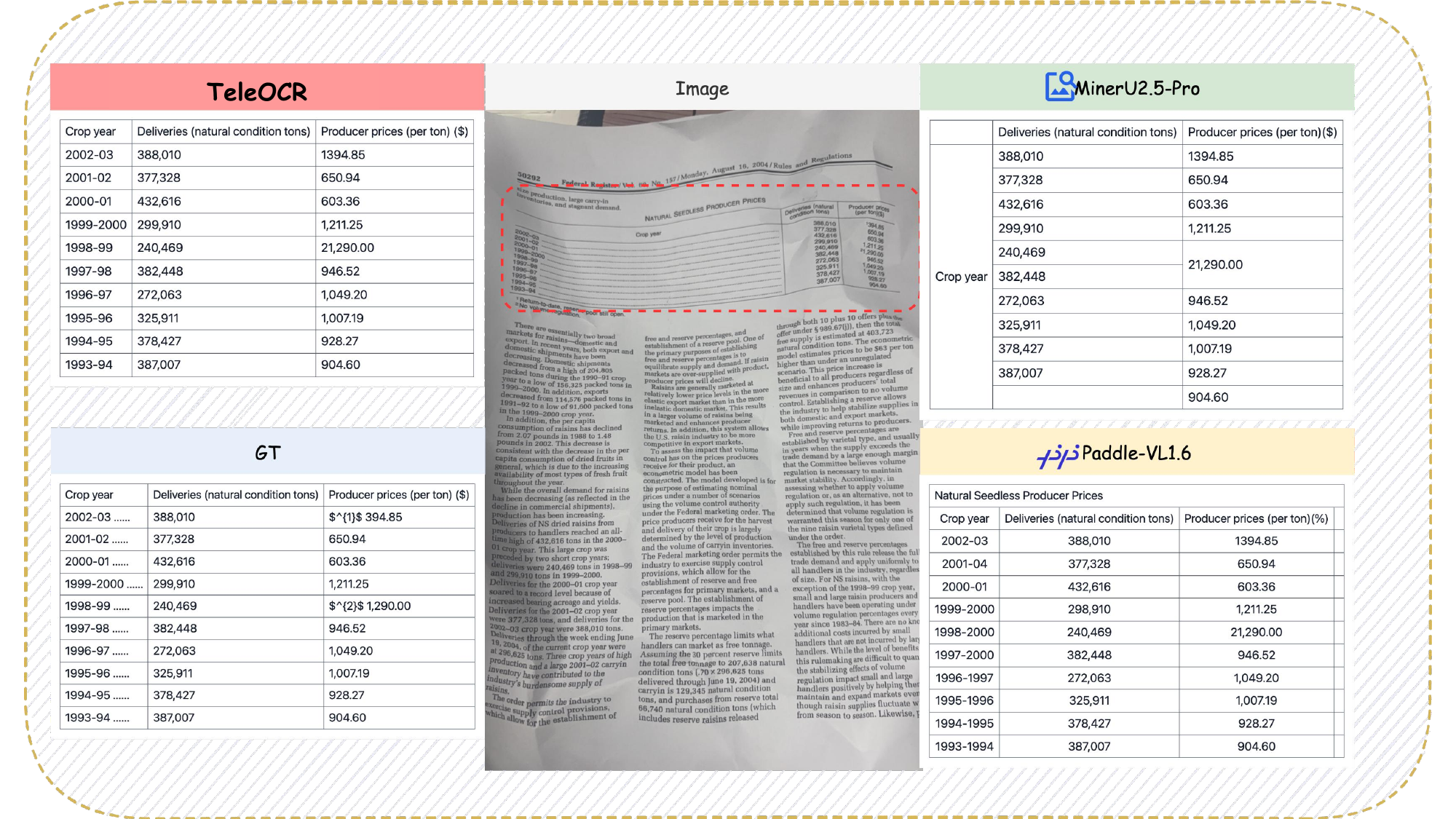}
\caption{Qualitative comparison on a small table embedded in a camera-captured newspaper page. TeleOCR accurately restores the three-column structure and numerical correspondences.}
\label{fig:table_2}
\end{figure}

The newspaper case in Figure \ref{fig:table_2} contains a small table embedded in dense text and affected by page curvature. Competing methods introduce incorrect row-spanning structures or miss columns, while TeleOCR accurately restores the three-column layout and preserves the correspondence among crop year, deliveries, and producer prices.

\begin{figure}[H]
\centering
\includegraphics[width=1\textwidth, trim=0mm 0mm 0mm 0mm, clip]{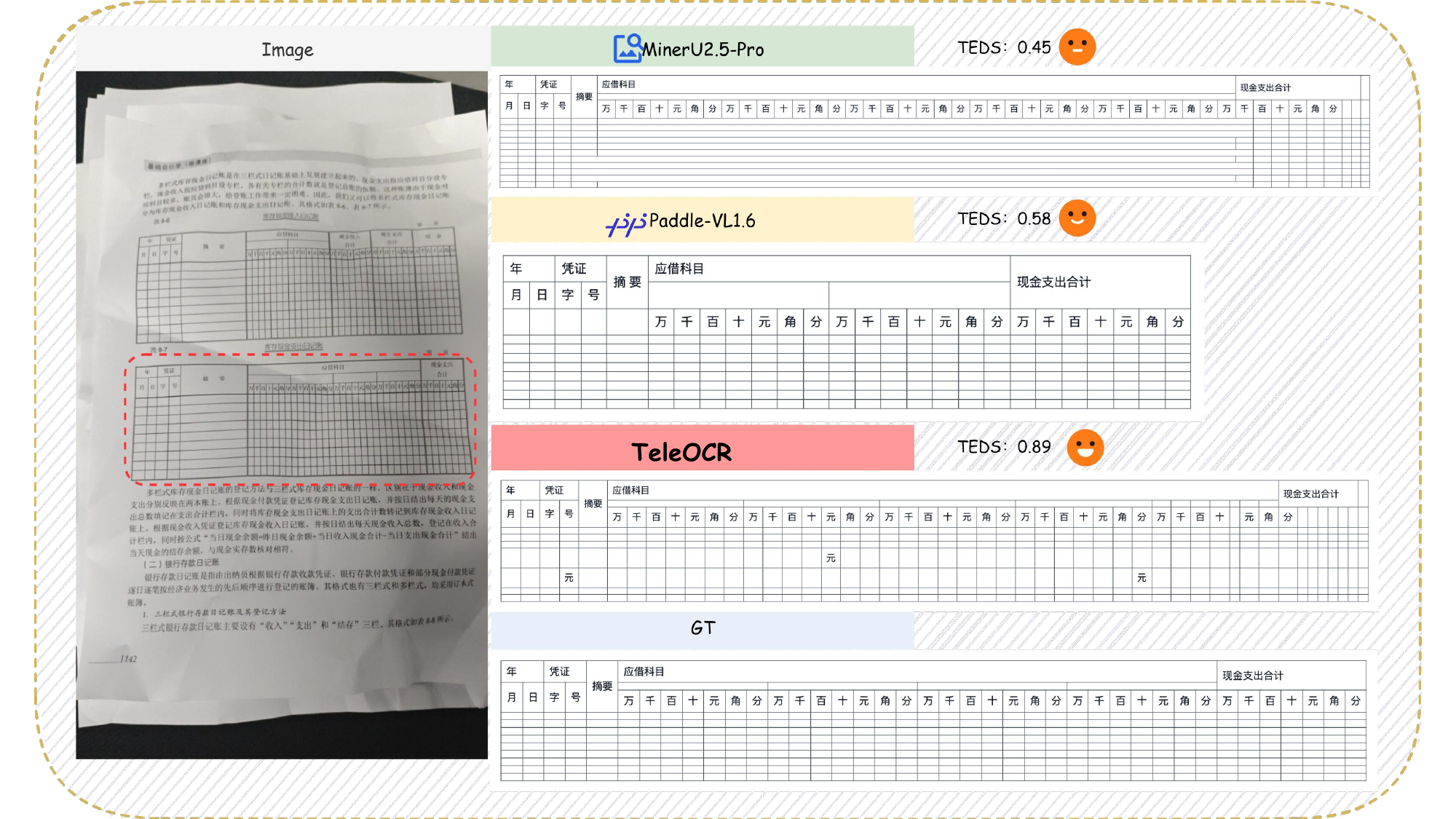}
\caption{Qualitative comparison on a dense financial ledger table. TeleOCR better preserves fine-grained grids, hierarchical headers, and empty cell structures.}
\label{fig:table_3}
\end{figure}

Figure \ref{fig:table_3} presents a dense financial ledger table with fine-grained grids and multi-level headers. This case requires accurate recovery of hierarchical headers, narrow columns, and empty cells. TeleOCR produces results closer to the GT, whereas MinerU2.5-Pro and Paddle-VL-1.6 tend to lose narrow columns, compress grids, or incorrectly merge cells. These results demonstrate the effectiveness of content-structure decoupled learning for table topology modeling.

\subsection{Formula Extraction }

We further compare formula extraction results under real-world degradations such as wrinkles, shadows, low resolution, and severe rotation. TeleOCR shows stronger robustness in preserving mathematical structures, including subscripts, superscripts, radicals, fractions, limits, and bracket scopes.



\begin{figure}[H]
\centering
\includegraphics[width=1\textwidth, trim=0mm 0mm 0mm 0mm, clip]{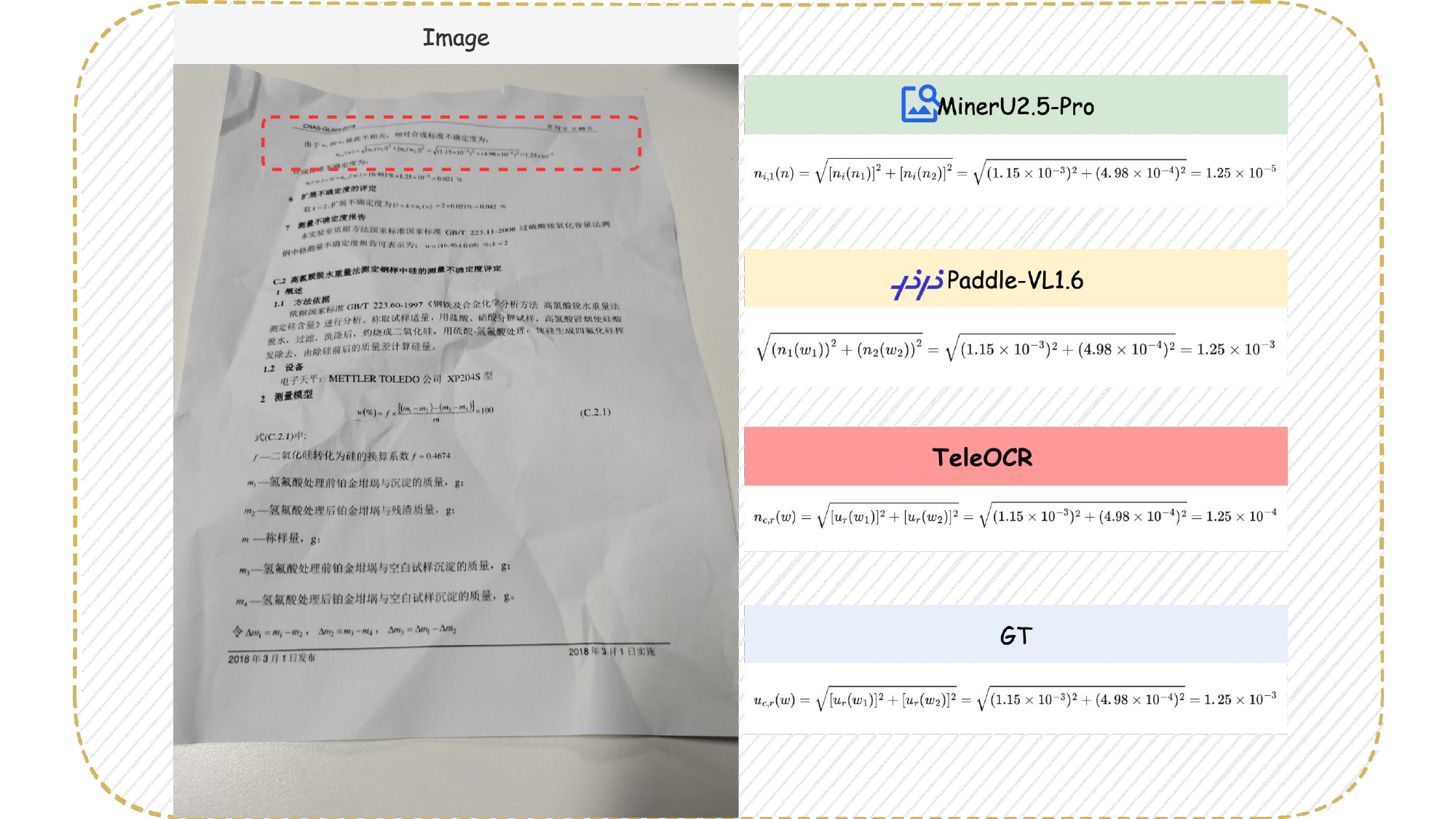}
\caption{Qualitative comparison on a low-resolution formula region with wrinkles and shadows. TeleOCR more accurately preserves radicals, subscripts, superscripts, and numerical expressions.}
\label{fig:formula_1}
\end{figure}

As shown in Figure \ref{fig:formula_1}, this case presents a challenging formula recognition scenario with paper wrinkles, local shadows, and a low-resolution formula region. MinerU2.5-Pro and Paddle-VL-1.6 can recognize major elements such as squares, radicals, and scientific notation, but they struggle with variable subscripts, exponent positions, and the final numerical magnitude. In comparison, TeleOCR more accurately preserves the summation relation inside the radical, the subscript/superscript structures, and variable symbols.

\begin{figure}[H]
\centering
\includegraphics[width=1\textwidth, trim=0mm 0mm 0mm 0mm, clip]{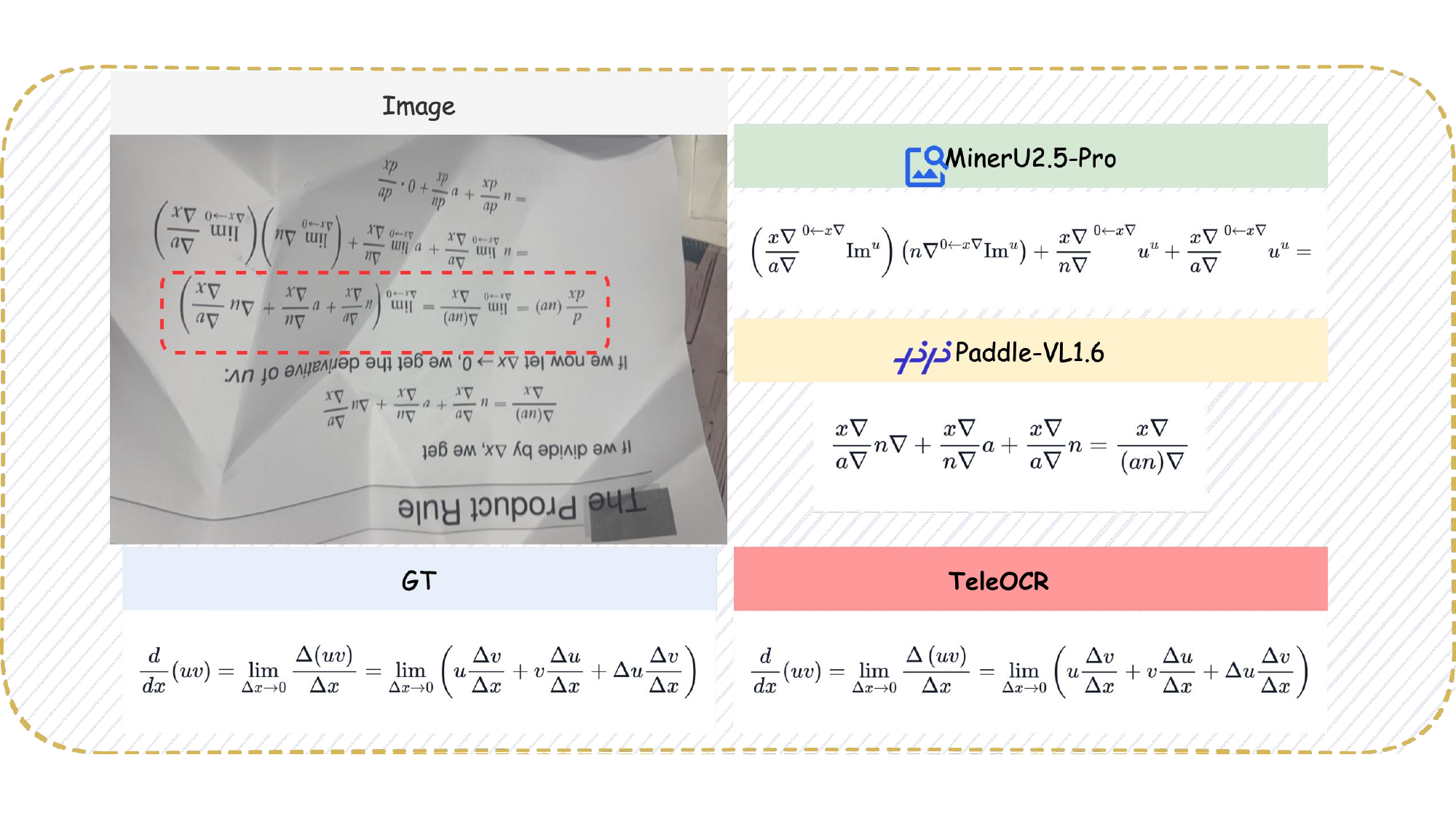}
\caption{Qualitative comparison on a low-resolution formula region with wrinkles and shadows. TeleOCR more accurately preserves radicals, subscripts, superscripts, and numerical expressions.}
\label{fig:formula_2}
\end{figure}

Figure \ref{fig:formula_2} further illustrates a more extreme camera-captured condition, where the page is severely rotated and the formula appears upside down. The example contains complex structures including limits, fractions, brackets, and product rule derivations. MinerU2.5-Pro detects several formula symbols but fails to recover the global orientation and structural relationships, while Paddle-VL-1.6 only reconstructs a partial formula fragment. In contrast, TeleOCR successfully recovers the complete formula expression under this upside-down condition, achieving high consistency with the GT. This illustrates the complementary benefits of deformation-aware learning and formula structure-aware decoupled learning for real-world camera-captured documents.

\section{Benchmark Evaluation Details}

We evaluate document parsing performance on OmniDocBench v1.6, Wild OmniDocBench v1.5, and PureDocBench. For consistency across benchmarks, Wild OmniDocBench v1.5 and PureDocBench are evaluated using the OmniDocBench-style end-to-end protocol. Model predictions are converted to Markdown files and matched against the benchmark ground truth using the \texttt{quick\_match} strategy in the OmniDocBench evaluator.

The evaluation covers four element groups: text blocks, display formulas, tables, and reading order. Text blocks and reading order are measured by normalized edit distance, display formulas are measured by both edit distance and CDM, and tables are measured by TEDS and edit distance. Following OmniDocBench v1.6, we report TextEdit, FormulaCDM, TableTEDS, TableTEDS-S, and ReadOrderEdit. The overall score is computed as:
\[
\mathrm{Overall} =
\frac{(1-\mathrm{TextEdit}) \times 100
+ \mathrm{FormulaCDM} \times 100
+ \mathrm{TableTEDS} \times 100}{3}.
\]
Here, lower TextEdit and ReadOrderEdit indicate better performance, while higher FormulaCDM, TableTEDS, TableTEDS-S, and Overall indicate better performance.

For all models compared on the same benchmark, we keep the evaluator version, matching strategy, metric set, and timeout settings fixed. To handle long or complex pages, the page matching timeout and truncated quick-match timeout are both set to 1200 seconds, with \texttt{timeout\_fallback\_max\_chunk\_span=200} and \texttt{timeout\_fallback\_order\_penalty=0.05}. For Wild OmniDocBench v1.5 and PureDocBench, we use the same metric and matching settings, replacing only the ground-truth file and prediction directory with the corresponding benchmark paths.

It is worth noting that PureDocBench is substantially more challenging, and TeleOCR exhibited severe repetitive generation on a small number of cases. For these samples, we removed the corresponding invalid Markdown prediction files before scoring. This does not affect evaluation fairness, because missing predictions are assigned a score of zero under the evaluation protocol. The removed files are listed in Table~\ref{tab:puredocbench_removed_predictions}. No predictions were removed from the digital-degraded subset.

\begin{table}[h]
\centering
\caption{Markdown prediction files removed from TeleOCR for PureDocBench evaluation.}
\label{tab:puredocbench_removed_predictions}
\resizebox{\linewidth}{!}{
\begin{tabular}{ll}
\hline
Subset & Removed prediction file \\
\hline
clean & \texttt{employee\_handbook\_001\_.md} \\
real\_degraded & \texttt{customs\_packing\_019\_\_Multi-Country-Re-Export-Trade-Customs-Documentation.md} \\
real\_degraded & \texttt{employee\_handbook\_002\_Manufacturing\_Safety.md} \\
real\_degraded & \texttt{itinerary\_020\_\_International\_Summit\_Schedule\_Overview.md} \\
real\_degraded & \texttt{professional\_cert\_018\_International\_PE\_Mutual\_Recognition\_-\_Color-Coded\_Zone\_Board.md} \\
real\_degraded & \texttt{slides\_006\_\_.md} \\
\hline
\end{tabular}
}
\end{table}

\section{Layout Visualization of Distorted Documents}
\begin{figure}[H]
\centering
\includegraphics[width=0.8\textwidth, trim=0mm 0mm 160mm 0mm, clip]{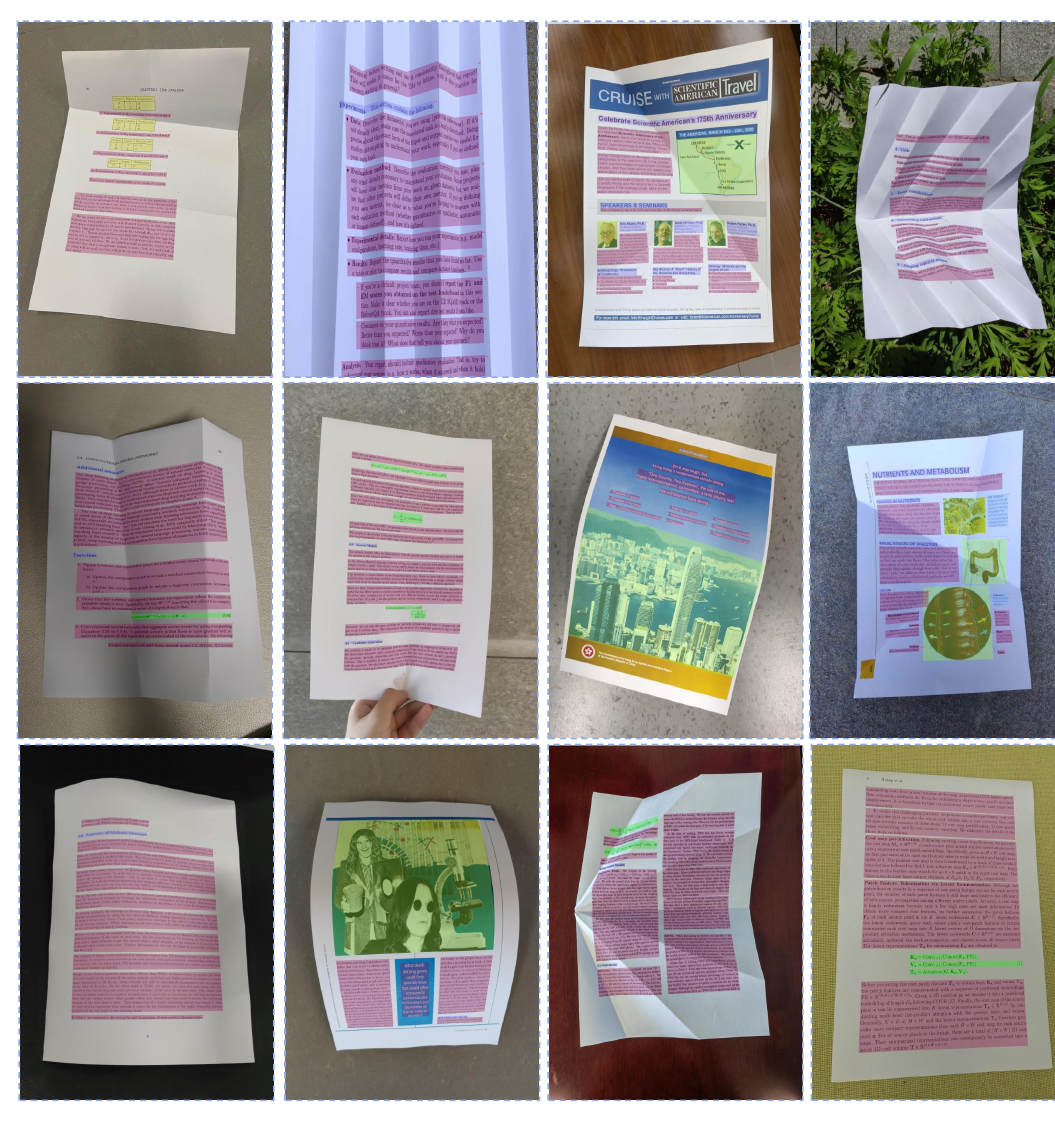} 
\caption{Parsing evaluation on the DIR300 dataset.}
\label{fig:dir300}
\end{figure}

\begin{figure}[H]
\centering
\includegraphics[width=0.8\textwidth, trim=0mm 50mm 45mm 0mm, clip]{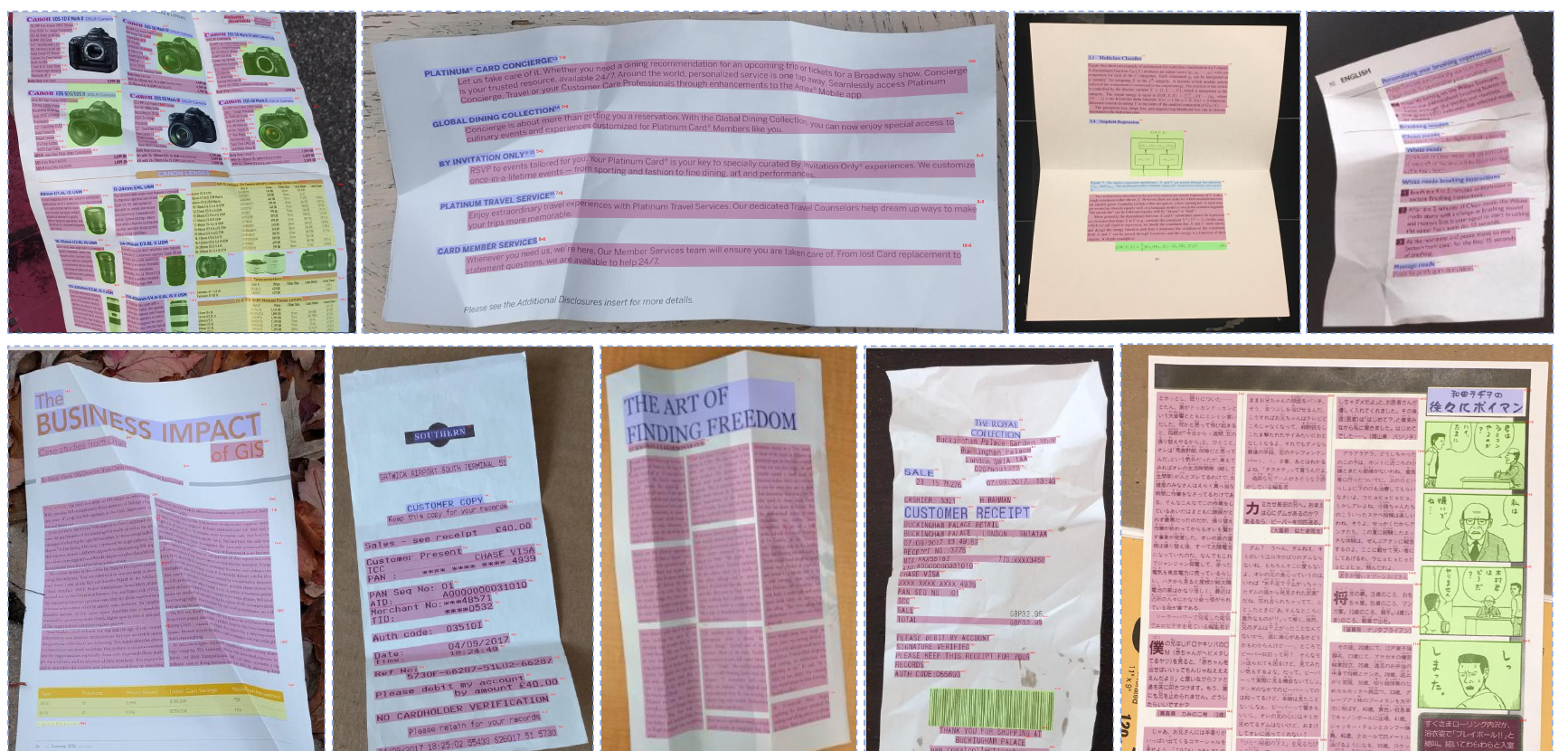} 
\caption{Parsing evaluation on the DocUNet dataset.}
\label{fig:docnnet}
\end{figure}

To evaluate the model's ability to understand complex document deformations, we conduct a visual evaluation on the public dewarping datasets DocUNet~\cite{ma2018docunet} and DIR300~\cite{feng2022geometric}, with representative results shown in Fig.~\ref{fig:dir300} and Fig.~\ref{fig:docnnet}. TeleOCR directly performs layout and content parsing on distorted documents without dewarping preprocessing or a dedicated rectification model, demonstrating robust parsing under complex geometric deformations.

\end{document}